# A Knowledge Compilation Map

**Adnan Darwiche** DARWICHE@CS.UCLA.EDU
*Computer Science Department*
*University of California, Los Angeles*
*Los Angeles, CA 90095, USA*

**Pierre Marquis** MARQUIS@CRIL.UNIV-ARTOIS.FR
*Université d'Artois*
*F-62307, Lens Cedex, France*

## Abstract

We propose a perspective on knowledge compilation which calls for analyzing different compilation approaches according to two key dimensions: the succinctness of the target compilation language, and the class of queries and transformations that the language supports in polytime. We then provide a knowledge compilation map, which analyzes a large number of existing target compilation languages according to their succinctness and their polytime transformations and queries. We argue that such analysis is necessary for placing new compilation approaches within the context of existing ones. We also go beyond classical, flat target compilation languages based on CNF and DNF, and consider a richer, nested class based on directed acyclic graphs (such as OBDDs), which we show to include a relatively large number of target compilation languages.

## 1. Introduction

Knowledge compilation has emerged recently as a key direction of research for dealing with the computational intractability of general propositional reasoning (Darwiche, 1999; Cadoli & Donini, 1997; Boufkhad, Grégoire, Marquis, Mazure, & Saïs, 1997; Khardon & Roth, 1997; Selman & Kautz, 1996; Schrag, 1996; Marquis, 1995; del Val, 1994; Dechter & Rish, 1994; Reiter & de Kleer, 1987). According to this direction, a propositional theory is compiled off-line into a target language, which is then used on-line to answer a large number of queries in polytime. The key motivation behind knowledge compilation is to push as much of the computational overhead into the off-line phase, which is amortized over all on-line queries. But knowledge compilation can serve other important purposes as well. For example, target compilation languages and their associated algorithms can be very simple, allowing one to develop on-line reasoning systems for simple software and hardware platforms. Moreover, the simplicity of algorithms that operate on compiled languages help in streamlining the effort of algorithmic design into a single task: that of generating the smallest compiled representations possible, as that turns out to be the main computational bottleneck in compilation approaches.

There are three key aspects of any knowledge compilation approach: the succinctness of the target language into which the propositional theory is compiled; the class of queries that can be answered in polytime based on the compiled representation; and the class of transformations that can be applied to the representation in polytime. The AI literature has thus far focused mostly on target compilation languages which are variations on DNF and CNF formulas, such as Horn theories and prime implicates. Moreover, it has focused mostly on clausal entailment queries, with very little discussion of tractable transformations on compiled theories.

The goal of this paper is to provide a broad perspective on knowledge compilation by considering a relatively large number of target compilation languages and analyzing them according to their *succinctness* and the class of *queries/transformations* that they admit in polytime.

Instead of focusing on classical, *flat* target compilation languages based on CNF and DNF, we consider a richer, *nested* class based on representing propositional sentences using directed acyclic graphs, which we refer to as NNF. We identify a number of target compilation languages that have been presented in the AI, formal verification, and computer science literature and show that they are special cases of NNF. For each such class, we list the extra conditions that need to be imposed on NNF to obtain the specific class, and then identify the set of queries and transformations that the class supports in polytime. We also provide cross-rankings of the different subsets of NNF, according to their succinctness and the polytime operations they support.

The main contribution of this paper is then a *map* for deciding the target compilation language that is most suitable for a particular application. Specifically, we propose that one starts by identifying the set of queries and transformations needed for their given application, and then choosing the most succinct language that supports these operations in polytime.

This paper is structured as follows. We start by formally defining the NNF language in Section 2, where we list a number of conditions on NNF that give rise to a variety of target compilation languages. We then study the succinctness of these languages in Section 3 and provide a cross-ranking that compares them according to this measure. We consider a number of queries and their applications in Section 4 and compare the different target compilation languages according to their tractability with respect to these queries. Section 5 is then dedicated to a class of transformations, their applications, and their tractability with respect to the different target compilation languages. We finally close in Section 6 by some concluding remarks. Proofs of all theorems are delegated to Appendix A.

## 2. The NNF Language

We consider more than a dozen languages in this paper, all of which are subsets of the NNF language, which is defined formally as follows (Darwiche, 1999, 2001a).

**Definition 2.1** *Let PS be a denumerable set of propositional variables. A sentence in* $\texttt{NNF}_{PS}$ *is a rooted, directed acyclic graph (DAG) where each leaf node is labeled with true, false,* $X$ *or* $\neg X$*,* $X \in PS$*; and each internal node is labeled with* $\wedge$ *or* $\vee$ *and can have arbitrarily many children. The size of a sentence* $\Sigma$ *in* $\texttt{NNF}_{PS}$*, denoted* $\mid \Sigma \mid$*, is the number of its DAG edges. Its height is the maximum number of edges from the root to some leaf in the DAG.*

Figure 1 depicts a sentence in NNF, which represents the odd parity function (we omit reference to variables PS when no confusion is anticipated). Any propositional sentence can be represented as a sentence in NNF, so the NNF language is *complete*.

It is important here to distinguish between a *representation* language and a *target compilation* language. A representation language is one which we expect humans to read and write with some ease. The language of CNF is a popular representation language, and so is the language of Horn clauses (especially when expressed in rules form). On other hand, a target compilation language does not need to be suitable for human specification and interpretation, but should be tractable enough to permit a non-trivial number of polytime queries and/or transformations. We will consider a number of target compilation languages that do not qualify as representation languages from this perspective, as they are not suitable for humans to construct or interpret. We will also consider a number of representation languages that do not qualify as target compilation languages.[1]

A formal characterization of representation languages is outside the scope of this paper. But for a language to qualify as a target compilation language, we will require that it permits a polytime clausal entailment test. Note that a polytime consistency test is not sufficient here, as only one consistency test on a given theory does not justify its compilation. Given this definition, NNF does

1. It appears that when proposing target compilation languages in the AI literature, there is usually an implicit requirement that the proposed language is also a representation language. As we shall see later, however, the most powerful target compilation languages are not suitable for humans to specify or interpret directly.

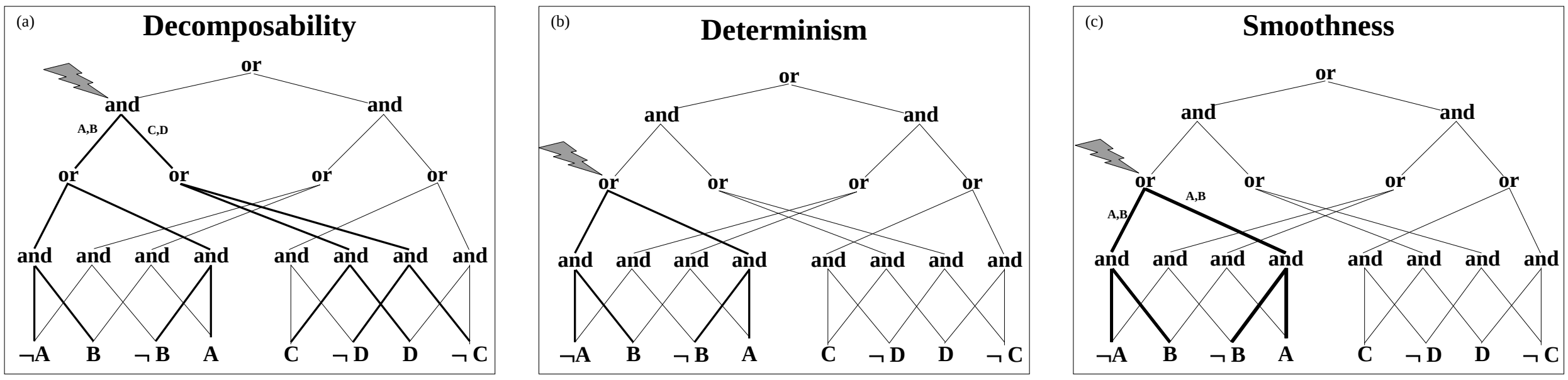

Figure 1: A sentence in NNF. Its size is 30 and height is 4.

not qualify as a target compilation language unless P=NP (Papadimitriou, 1994), but many of its subsets do. We define a number of these subsets below, each of which is obtained by imposing further conditions on NNF.

We will distinguish between two key subsets of NNF: *flat* and *nested* subsets. We first consider flat subsets, which result from imposing combinations of the following properties:

- **Flatness:** The height of each sentence is at most 2. The sentence in Figure 3 is flat, but the one in Figure 1 is not.
- **Simple-disjunction:** The children of each or-node are leaves that share no variables (the node is a *clause*).
- **Simple-conjunction:** The children of each and-node are leaves that share no variables (the node is a *term*). The sentence in Figure 3 satisfies this property.

**Definition 2.2** *The language* f-NNF *is the subset of* NNF *satisfying flatness. The language* CNF *is the subset of* f-NNF *satisfying simple–disjunction. The language* DNF *is the subset of* f-NNF *satisfying simple–conjunction.*

CNF does not permit a polytime clausal entailment test (unless P=NP) and, hence, does not qualify as a target compilation language. But its dual DNF does.

The following subset of CNF, *prime implicates,* has been quite influential in computer science:

**Definition 2.3** *The language* PI *is the subset of* CNF *in which each clause entailed by the sentence is subsumed by a clause that appears in the sentence; and no clause in the sentence is subsumed by another.*

A dual of PI, *prime implicants* IP, can also be defined.

**Definition 2.4** *The language* IP *is the subset of* DNF *in which each term entailing the sentence subsumes some term that appears in the sentence; and no term in the sentence is subsumed by another term.*

There has been some work on representing the set of prime implicates of a propositional theory in a compact way, allowing an exponential number of prime implicates to be represented in polynomial space in certain cases—see for example the TRIE representation in (de Kleer, 1992), the ZBDD representation used in (Simon & del Val, 2001), and the implicit representation based on meta-products, as proposed in (Madre & Coudert, 1992). These representations are different from the language PI in the sense that they do not necessarily support the same queries and transformations that we report in Tables 5 and 7. They also exhibit different succinctness relationships than the ones we report in Table 3.

Horn theories (and renamable Horn theories) represent another target compilation subset of CNF, but we do not consider it here since we restrict our attention to *complete languages* **L** only, i.e., we require that every propositional sentence is logically equivalent to an element of **L**.

We now consider *nested* subsets of the NNF language, which do not impose any restriction on the height of a sentence. Instead, these subsets result from imposing one or more of the following conditions: *decomposability, determinism, smoothness, decision,* and *ordering.* We start by defining the first three properties. From here on, if $C$ is a node in an NNF, then $Vars(C)$ denotes the set of all variables that label the descendants of node $C$. Moreover, if $\Sigma$ is an NNF sentence rooted at $C$, then $Vars(\Sigma)$ is defined as $Vars(C)$.

- **Decomposability** (Darwiche, 1999, 2001a). An NNF satisfies this property if for each conjunction $C$ in the NNF, the conjuncts of $C$ do not share variables. That is, if $C_1, \ldots, C_n$ are the children of and-node $C$, then $Vars(C_i) \cap Vars(C_j) = \emptyset$ for $i \neq j$. Consider the and-node marked in Figure 1(a). This node has two children, the first contains variables $A, B$ while the second contains variables $C, D$. This and-node is then decomposable since the two children do not share variables. Each other and-node in Figure 1(a) is also decomposable and, hence, the NNF in this figure is decomposable.

- **Determinism** (Darwiche, 2001b): An NNF satisfies this property if for each disjunction $C$ in the NNF, each two disjuncts of $C$ are logically contradictory. That is, if $C_1, \ldots, C_n$ are the children of or-node $C$, then $C_i \wedge C_j \models \mathit{false}$ for $i \neq j$. Consider the or-node marked in Figure 1(b), which has two children corresponding to sub-sentences $\neg A \wedge B$ and $\neg B \wedge A$. The conjunction of these two sub-sentences is logically contradictory. The or-node is then deterministic and so are the other or-nodes in Figure 1(b). Hence, the NNF in this figure is deterministic.

- **Smoothness** (Darwiche, 2001b): An NNF satisfies this property if for each disjunction $C$ in the NNF, each disjunct of $C$ mentions the same variables. That is, if $C_1, \ldots, C_n$ are the children of or-node $C$, then $Vars(C_i) = Vars(C_j)$ for $i \neq j$. Consider the marked or-node in Figure 1(c). This node has two children, each of which mentions variables $A, B$. This or-node is then smooth and so are the other or-nodes in Figure 1(c). Hence, the NNF in this figure is smooth.

It is hard to ensure decomposability. It is also hard to ensure determinism while preserving decomposability. Yet any sentence in NNF can be smoothed in polytime, while preserving decomposability and determinism. Preserving flatness, however, may blow-up the size of given NNF. Hence, smoothness is not that important from a complexity viewpoint unless we have flatness.

The properties of decomposability, determinism and smoothness lead to a number of interesting subsets of NNF.

**Definition 2.5** *The language* DNNF *is the subset of* NNF *satisfying decomposability;* d-NNF *is the subset satisfying determinism;* s-NNF *is the subset satisfying smoothness;* d-DNNF *is the subset satisfying decomposability and determinism; and* sd-DNNF *is the subset satisfying decomposability, determinism and smoothness.*

Note that DNF is a strict subset of DNNF (Darwiche, 1999, 2001a). The following *decision* property comes from the literature on *binary decision diagrams* (Bryant, 1986).

**Definition 2.6 (Decision)** *A decision node $N$ in an* NNF *sentence is one which is labeled with true, false, or is an or-node having the form $(X \wedge \alpha) \vee (\neg X \wedge \beta)$, where $X$ is a variable, $\alpha$ and $\beta$ are decision nodes. In the latter case, $dVar(N)$ denotes the variable $X$.*

**Definition 2.7** *The language* BDD *is the set of* NNF *sentences, where the root of each sentence is a decision node.*

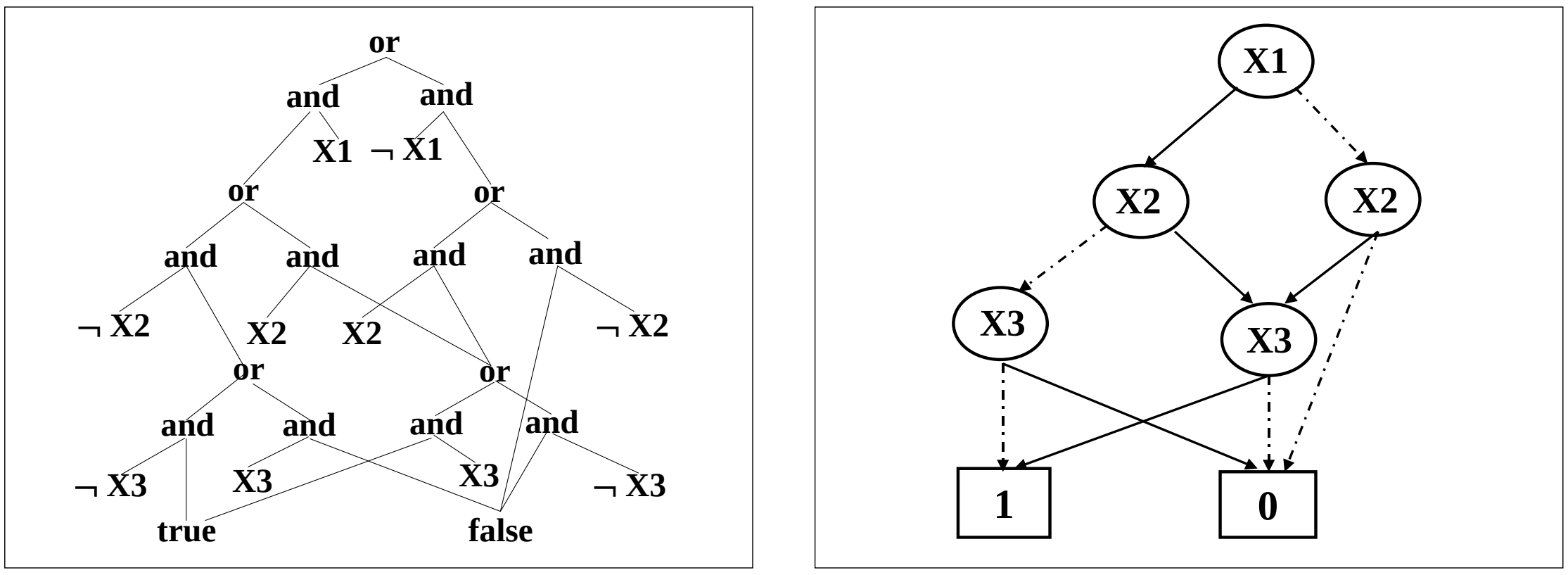

Figure 2: On the left, a sentence in the BDD language. On the right, its corresponding binary decision diagram.

The NNF sentence in Figure 2 belongs to the BDD subset.

The BDD language corresponds to *binary decision diagrams (BDDs),* as known in the formal verification literature (Bryant, 1986). Binary decision diagrams are depicted using a more compact notation though: the labels *true* and *false* are denoted by 1 and 0, respectively; and each decision node

is denoted by

. The BDD sentence on the left of Figure 2 corresponds to the binary decision diagram on the right of Figure 2. Obviously enough, every NNF sentence that satisfies the decision property is also deterministic. Therefore, BDD is a subset of d-NNF.

As we show later, BDD does not qualify as a target compilation language (unless P=NP), but the following subset does.

**Definition 2.8** FBDD *is the intersection of* DNNF *and* BDD*.*

That is, each sentence in FBDD is decomposable and satisfies the decision property. The FBDD language corresponds to *free binary decision diagrams (FBDDs),* as known in formal verification (Gergov & Meinel, 1994a). An FBDD is usually defined as a BDD that satisfies the *read-once property:* on each path from the root to a leaf, a variable can appear at most once. FBDDs are also known as read-once branching programs in the theory literature. Imposing the read-once property on a BDD is equivalent to imposing the decomposability property on its corresponding BDD sentence.

A more influential subset of the BDD language is obtained by imposing the *ordering* property:

**Definition 2.9 (Ordering)** *Let $<$ be a total ordering on the variables PS. The language* $\texttt{OBDD}_<$ *is the subset of* FBDD *satisfying the following property: if $N$ and $M$ are or-nodes, and if $N$ is an ancestor of node $M$, then $dVar(N) < dVar(M)$.*

**Definition 2.10** *The language* OBDD *is the union of all* $\texttt{OBDD}_<$ *languages.*

The OBDD language corresponds to the well–known *ordered binary decision diagrams (OBDDs)* (Bryant, 1986).

Our final language definition is as follows:

**Definition 2.11** MODS *is the subset of* DNF *where every sentence satisfies determinism and smoothness.*

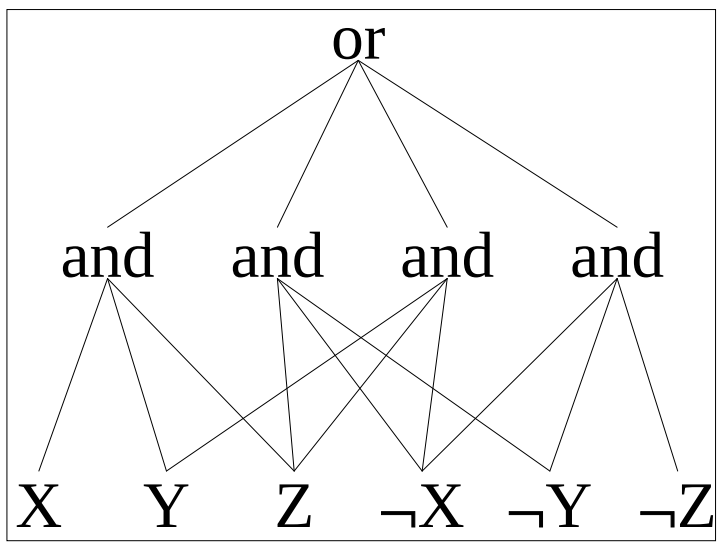

Figure 3: A sentence in language MODS.

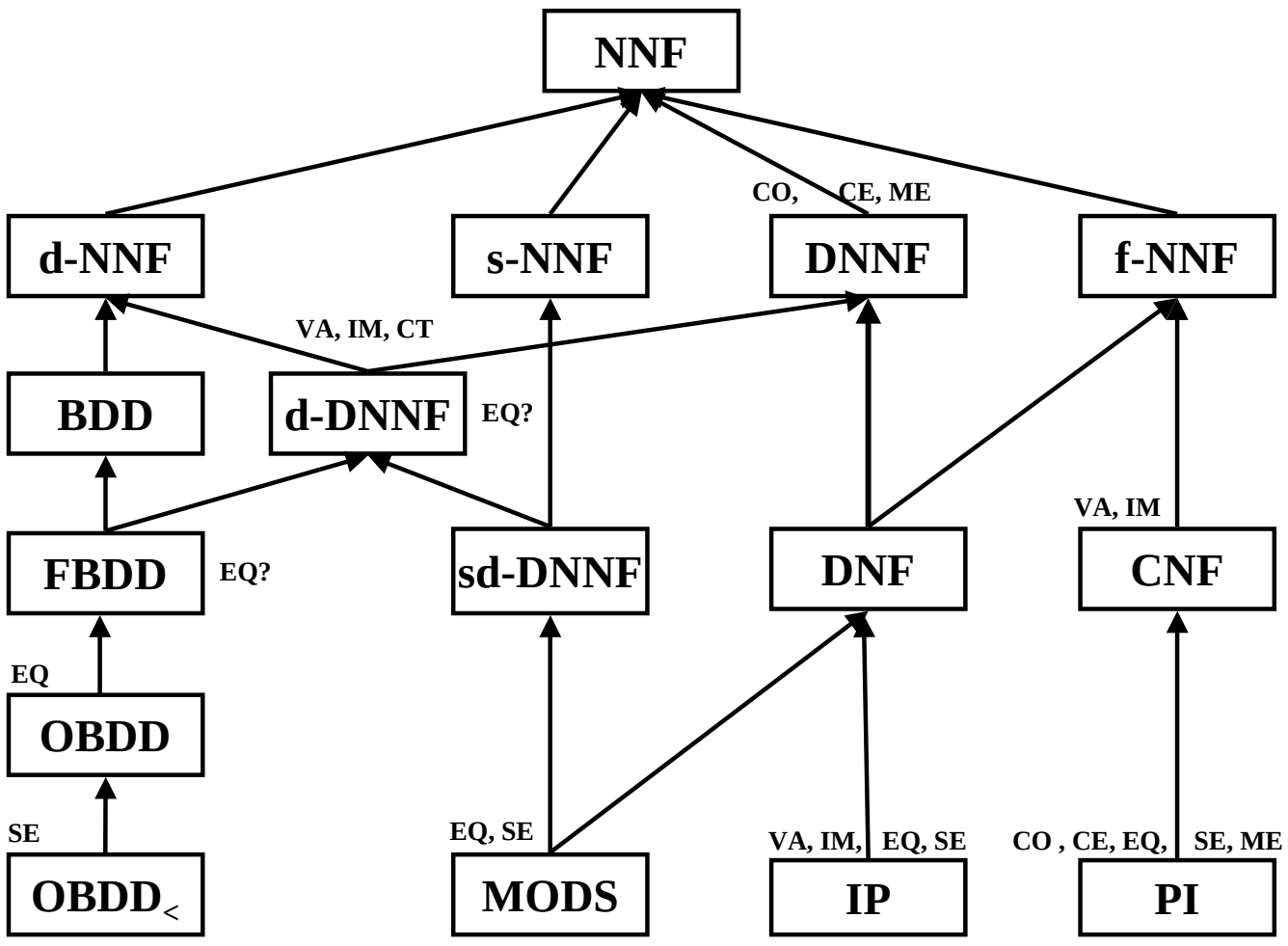

Figure 4: The set of DAG-based languages considered in this paper. An edge $\mathbf{L}_1 \rightarrow \mathbf{L}_2$ means that $\mathbf{L}_1$ is a proper subset of $\mathbf{L}_2$. Next to each subset, we list the polytime queries supported by the subset but not by any of its ancestors (see Section 4).

Figure 3 depicts a sentence in MODS. As we show later, MODS is the most tractable NNF subset we shall consider (together with $\text{OBDD}_<$). This is not surprising since from the syntax of a sentence in MODS, one can immediately recover the sentence models.

The languages we have discussed so far are depicted in Figure 4, where arrows denote set inclusion. Table 1 lists the acronyms of all of these languages, together with their descriptions. Table 2 lists the key language properties discussed in this section, together with a short description of each.

## 3. On the Succinctness of Compiled Theories

We have discussed more than a dozen subsets of the NNF language. Some of these subsets are well known and have been studied extensively in the computer science literature. Others, such as DNNF (Darwiche, 2001a, 1999) and d-DNNF (Darwiche, 2001b), are relatively new. The question now is: What subset should one adopt for a particular application? As we argue in this paper, that depends on three key properties of the language: its succinctness, the class of tractable queries it supports, and the class of tractable transformations it admits.

| **Acronym** | **Description** |
|---|---|
| NNF | Negation Normal Form |
| DNNF | Decomposable Negation Normal Form |
| d-NNF | Deterministic Negation Normal Form |
| s-NNF | Smooth Negation Normal Form |
| f-NNF | Flat Negation Normal Form |
| d-DNNF | Deterministic Decomposable Negation Normal Form |
| sd-DNNF | Smooth Deterministic Decomposable Negation Normal Form |
| BDD | Binary Decision Diagram |
| FBDD | Free Binary Decision Diagram |
| OBDD | Ordered Binary Decision Diagram |
| $\texttt{OBDD}_<$ | Ordered Binary Decision Diagram (using order $<$) |
| DNF | Disjunctive Normal Form |
| CNF | Conjunctive Normal Form |
| PI | Prime Implicates |
| IP | Prime Implicants |
| MODS | Models |

Table 1: Language acronyms.

| **Property** | **Short Description** |
|---|---|
| Flatness | The height of NNF is at most 2 |
| Simple Disjunction | Every disjunction is a clause, where literals share no variables |
| Simple Conjunction | Every conjunction is a term, where literals share no variables |
| Decomposability | Conjuncts do not share variables |
| Determinism | Disjuncts are logically disjoint |
| Smoothness | Disjuncts mention the same set of variables |
| Decision | A node of the form *true*, *false*, or $(X \wedge \alpha \vee \neg X \wedge \beta)$, where $X$ is a variable and $\alpha, \beta$ are decision nodes |
| Ordering | Decision variables appear in the same order on any path in the NNF |

Table 2: Language properties.

Our goal in this and the following sections is to construct a map on which we place different subsets of the NNF language according to the above criteria. This map will then serve as a guide to system designers in choosing the target compilation language most suitable to their application. It also provides an example paradigm for studying and evaluating further target compilation languages. We start with a study of succinctness[2] in this section (Gogic, Kautz, Papadimitriou, & Selman, 1995).

**Definition 3.1 (Succinctness)** *Let* $\mathbf{L}_1$ *and* $\mathbf{L}_2$ *be two subsets of* NNF. $\mathbf{L}_1$ *is at least as succinct as* $\mathbf{L}_2$*, denoted* $\mathbf{L}_1 \leq \mathbf{L}_2$*, iff there exists a polynomial* $p$ *such that for every sentence* $\alpha \in \mathbf{L}_2$*, there exists an equivalent sentence* $\beta \in \mathbf{L}_1$ *where* $|\beta| \leq p(|\alpha|)$*. Here,* $|\alpha|$ *and* $|\beta|$ *are the sizes of* $\alpha$ *and* $\beta$*, respectively.*

We stress here that we do not require that there exists a function that computes $\beta$ given $\alpha$ in *polytime*; we only require that a *polysize* $\beta$ exists. Yet, our proofs in Appendix A contain specific algorithms for computing $\beta$ from $\alpha$ in certain cases. The relation $\leq$ is clearly reflexive and transitive, hence, a pre-ordering. One can also define the relation $<$, where $\mathbf{L}_1 < \mathbf{L}_2$ iff $\mathbf{L}_1 \leq \mathbf{L}_2$ and $\mathbf{L}_2 \not\leq \mathbf{L}_1$.

**Proposition 3.1** *The results in Table 3 hold.*

An occurrence of $\leq$ in the cell of row $r$ and column $c$ of Table 3 means that the fragment $\mathbf{L}_r$ given at row $r$ is at least as succinct as the fragment $\mathbf{L}_c$ given at column $c$. An occurrence of $\not\leq$ (or $\not\leq^*$) means that $\mathbf{L}_r$ is not at least as succinct as $\mathbf{L}_c$ (provided that the polynomial hierarchy does not collapse in the case of $\not\leq^*$). Finally, the presence of a question mark reflects our ignorance about whether $\mathbf{L}_r$ is at least as succinct as $\mathbf{L}_c$. Figure 5 summarizes the results of Proposition 3.1 in terms of a directed acyclic graph.

A classical result in knowledge compilation states that it is not possible to compile any propositional formula $\alpha$ into a polysize data structure $\beta$ such that: $\alpha$ and $\beta$ entail the same set of clauses, and clausal entailment on $\beta$ can be decided in time polynomial in its size, unless $\mathsf{NP} \subseteq \mathsf{P/poly}$ (Selman & Kautz, 1996; Cadoli & Donini, 1997). This last assumption implies the collapse of the polynomial hierarchy at the second level (Karp & Lipton, 1980), which is considered very unlikely. We use this classical result from knowledge compilation in some of our proofs of Proposition 3.1, which explains why some of its parts are conditioned on the polynomial hierarchy not collapsing.

We have excluded the subsets BDD, s-NNF, d-NNF and f-NNF from Table 3 since they do not qualify as target compilation languages (see Section 4). We kept NNF and CNF though given their importance. Consider Figure 5 which depicts Table 3 graphically. With the exception of NNF and CNF, all other languages depicted in Figure 5 qualify as target compilation languages. Moreover, with the exception of language PI, DNNF is the most succinct among all target compilation languages—we know that PI is not more succinct than DNNF, but we do not know whether DNNF is more succinct than PI.

In between DNNF and MODS, there is a succinctness ordering of target compilation languages:

$$\texttt{DNNF} < \texttt{d-DNNF} < \texttt{FBDD} < \texttt{OBDD} < \texttt{OBDD}_< < \texttt{MODS}.$$

DNNF is obtained by imposing decomposability on NNF; d-DNNF by adding determinism; FBDD by adding decision; and OBDD and $\texttt{OBDD}_<$ by adding ordering (w.r.t. any total ordering on PS in the first case and a specific one in the second case). *Adding each of these properties reduces language succinctness (assuming that the polynomial hierarchy does not collapse).*

One important fact to stress here is that adding smoothness to d-DNNF does not affect its succinctness: the sd-DNNF and d-DNNF languages are equally succinct. It is also interesting to compare

2. A more general notion of space efficiency (model preservation for polysize reductions) exists (Cadoli, Donini, Liberatore, & Schaerf, 1996), but we do not need its full generality here.

| $\mathbf{L}$ | NNF | DNNF | d-DNNF | sd-DNNF | FBDD | OBDD | $\texttt{OBDD}_<$ | DNF | CNF | PI | IP | MODS |
|---|---|---|---|---|---|---|---|---|---|---|---|---|
| NNF | $\leq$ | $\leq$ | $\leq$ | $\leq$ | $\leq$ | $\leq$ | $\leq$ | $\leq$ | $\leq$ | $\leq$ | $\leq$ | $\leq$ |
| DNNF | $\not\leq^*$ | $\leq$ | $\leq$ | $\leq$ | $\leq$ | $\leq$ | $\leq$ | $\leq$ | $\not\leq^*$ | **?** | $\leq$ | $\leq$ |
| d-DNNF | $\not\leq^*$ | $\not\leq^*$ | $\leq$ | $\leq$ | $\leq$ | $\leq$ | $\leq$ | $\not\leq^*$ | $\not\leq^*$ | **?** | **?** | $\leq$ |
| sd-DNNF | $\not\leq^*$ | $\not\leq^*$ | $\leq$ | $\leq$ | $\leq$ | $\leq$ | $\leq$ | $\not\leq^*$ | $\not\leq^*$ | **?** | **?** | $\leq$ |
| FBDD | $\not\leq$ | $\not\leq$ | $\not\leq$ | $\not\leq$ | $\leq$ | $\leq$ | $\leq$ | $\not\leq$ | $\not\leq$ | $\not\leq$ | $\not\leq$ | $\leq$ |
| OBDD | $\not\leq$ | $\not\leq$ | $\not\leq$ | $\not\leq$ | $\not\leq$ | $\leq$ | $\leq$ | $\not\leq$ | $\not\leq$ | $\not\leq$ | $\not\leq$ | $\leq$ |
| $\texttt{OBDD}_<$ | $\not\leq$ | $\not\leq$ | $\not\leq$ | $\not\leq$ | $\not\leq$ | $\not\leq$ | $\leq$ | $\not\leq$ | $\not\leq$ | $\not\leq$ | $\not\leq$ | $\leq$ |
| DNF | $\not\leq$ | $\not\leq$ | $\not\leq$ | $\not\leq$ | $\not\leq$ | $\not\leq$ | $\not\leq$ | $\leq$ | $\not\leq$ | $\not\leq$ | $\leq$ | $\leq$ |
| CNF | $\not\leq$ | $\not\leq$ | $\not\leq$ | $\not\leq$ | $\not\leq$ | $\not\leq$ | $\not\leq$ | $\not\leq$ | $\leq$ | $\leq$ | $\not\leq$ | $\leq$ |
| PI | $\not\leq$ | $\not\leq$ | $\not\leq$ | $\not\leq$ | $\not\leq$ | $\not\leq$ | $\not\leq$ | $\not\leq$ | $\not\leq$ | $\leq$ | $\not\leq$ | **?** |
| IP | $\not\leq$ | $\not\leq$ | $\not\leq$ | $\not\leq$ | $\not\leq$ | $\not\leq$ | $\not\leq$ | $\not\leq$ | $\not\leq$ | $\not\leq$ | $\leq$ | $\leq$ |
| MODS | $\not\leq$ | $\not\leq$ | $\not\leq$ | $\not\leq$ | $\not\leq$ | $\not\leq$ | $\not\leq$ | $\not\leq$ | $\not\leq$ | $\not\leq$ | $\not\leq$ | $\leq$ |

Table 3: Succinctness of target compilation languages. $*$ means that the result holds unless the polynomial hierarchy collapses.

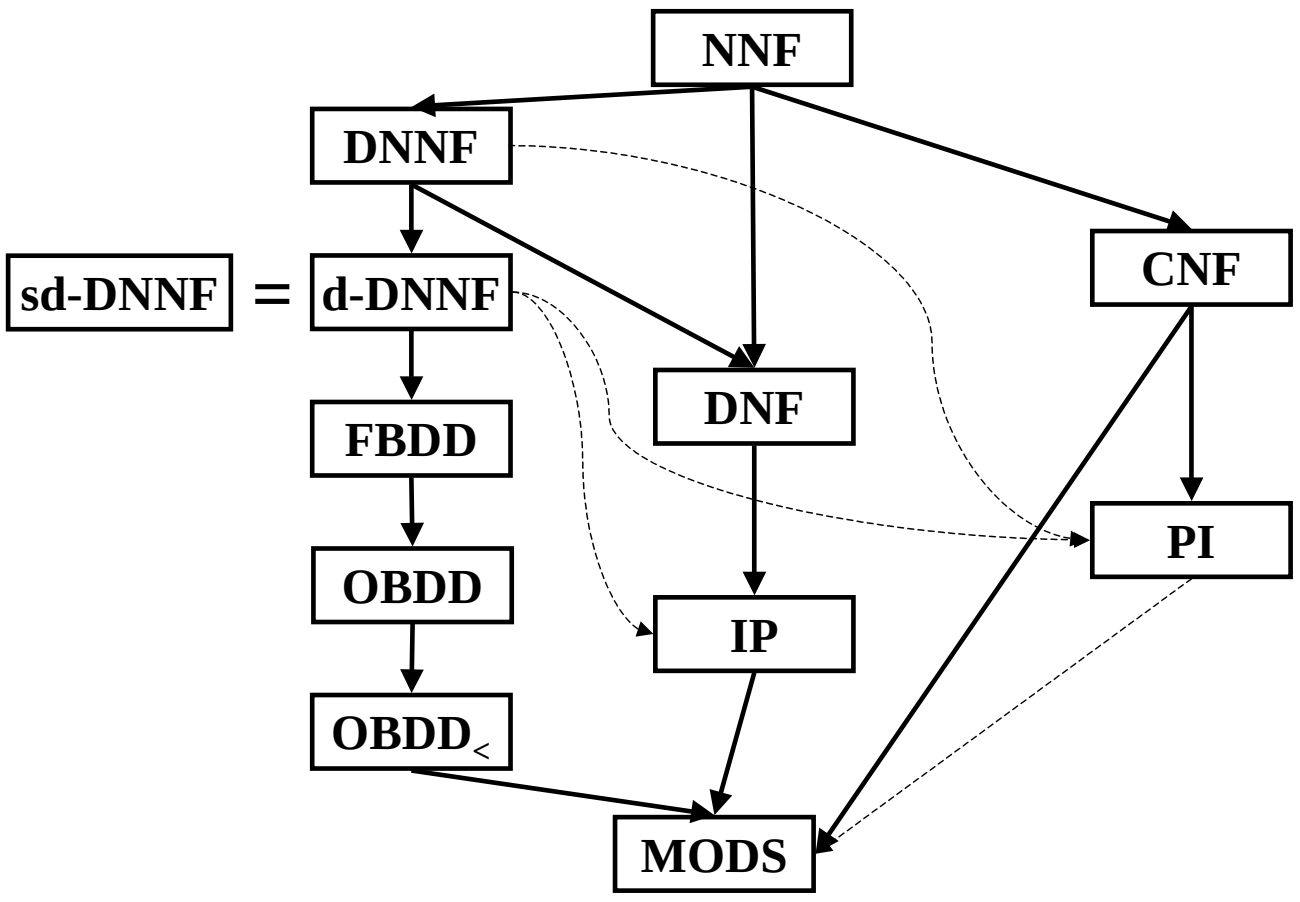

Figure 5: An edge $\mathbf{L}_1 \rightarrow \mathbf{L}_2$ indicates that $\mathbf{L}_1$ is strictly more succinct than $\mathbf{L}_2$: $\mathbf{L}_1 < \mathbf{L}_2$, while $\mathbf{L}_1 = \mathbf{L}_2$ indicates that $\mathbf{L}_1$ and $\mathbf{L}_2$ are equally succinct: $\mathbf{L}_1 \leq \mathbf{L}_2$ and $\mathbf{L}_2 \leq \mathbf{L}_1$. Dotted arrows indicate unknown relationships; for instance, the dotted arrow from DNNF to PI means that we do not know whether DNNF is at least as succinct as PI. Some of the edges are conditioned on the polynomial hierarchy not collapsing—see Table 3.

sd-DNNF (which is more succinct than the influential FBDD, OBDD and $\texttt{OBDD}_<$ languages) with MODS, which is a most tractable language. Both sd-DNNF and MODS are smooth, deterministic and decomposable. MODS, however, is flat and obtains its decomposability from the stronger condition of simple-conjunction. Therefore, sd-DNNF can be viewed as the result of relaxing from MODS the flatness and simple-conjunction conditions, while maintaining decomposability, determinism and smoothness. Relaxing these conditions moves the language three levels up the succinctness hierarchy, although it compromises only the polytime test for sentential entailment and possibly the one for equivalence as we show in Section 4.

## 4. Querying a Compiled Theory

In evaluating the suitability of a target compilation language to a particular application, the succinctness of the language must be balanced against the set of queries and transformations that it supports in polytime. We consider in this section a number of queries, each of which returns valuable information about a propositional theory, and then identify target compilation languages which provide polytime algorithms for answering such queries. We restrict our attention in this paper to the *existence* of polytime algorithms for answering queries, but we do not present the algorithms themselves. The interested reader is referred to (Darwiche, 2001a, 2001b, 1999; Bryant, 1986) for some of these algorithms and to the proofs of theorems in Appendix A for others.

The queries we consider are tests for consistency, validity, implicates (clausal entailment), implicants, equivalence, and sentential entailment. We also consider counting and enumerating theory models; see Table 4. One can also consider computing the probability of a propositional sentence, assuming that all variables are probabilistically independent. For the subsets we consider, however, this can be done in polytime whenever models can be counted in polytime.

From here on, **L** denotes a subset of language `NNF`.

**Definition 4.1 (CO, VA)** **L** *satisfies* **CO** *(***VA***) iff there exists a polytime algorithm that maps every formula* $\Sigma$ *from* **L** *to* 1 *if* $\Sigma$ *is consistent (valid), and to* 0 *otherwise.*

One of the main applications of compiling a theory is to enhance the efficiency of answering clausal entailment queries:

**Definition 4.2 (CE)** **L** *satisfies* **CE** *iff there exists a polytime algorithm that maps every formula* $\Sigma$ *from* **L** *and every clause* $\gamma$ *from* `NNF` *to* 1 *if* $\Sigma \models \gamma$ *holds, and to* 0 *otherwise.*

A key application of clausal entailment is in testing equivalence. Specifically, suppose we have a design expressed as a set of clauses $\Delta^d = \bigwedge_i \alpha_i$ and a specification expressed also as a set of clauses $\Delta^s = \bigwedge_j \beta_j$, and we want to test whether the design and specification are equivalent. By compiling each of $\Delta^d$ and $\Delta^s$ to targets $\Gamma^d$ and $\Gamma^s$ that support a polytime clausal entailment test, we can test the equivalence of $\Delta^d$ and $\Delta^s$ in polytime. That is, $\Delta^d$ and $\Delta^s$ are equivalent iff $\Gamma^d \models \beta_j$ for all $j$ and $\Gamma^s \models \alpha_i$ for all $i$.

A number of the target compilation languages we shall consider support a direct polytime equivalent test:

**Definition 4.3 (EQ, SE)** **L** *satisfies* **EQ** *(***SE***) iff there exists a polytime algorithm that maps every pair of formulas* $\Sigma$*,* $\Phi$ *from* **L** *to* 1 *if* $\Sigma \equiv \Phi$ *(*$\Sigma \models \Phi$*) holds, and to* 0 *otherwise.*

Note that sentential entailment (**SE**) is stronger than clausal entailment and equivalence. Therefore, if a language **L** satisfies **SE**, it also satisfies **CE** and **EQ**.

For completeness, we consider the following dual to **CE**:

**Definition 4.4 (IM)** **L** *satisfies* **IM** *iff there exists a polytime algorithm that maps every formula* $\Sigma$ *from* **L** *and every term* $\gamma$ *from* `NNF` *to* 1 *if* $\gamma \models \Sigma$ *holds, and to* 0 *otherwise.*

Finally, we consider counting and enumerating models:

**Definition 4.5 (CT)** **L** *satisfies* **CT** *iff there exists a polytime algorithm that maps every formula* $\Sigma$ *from* **L** *to a nonnegative integer that represents the number of models of* $\Sigma$ *(in binary notation).*

**Definition 4.6 (ME)** **L** *satisfies* **ME** *iff there exists a polynomial* $p(.,.)$ *and an algorithm that outputs all models of an arbitrary formula* $\Sigma$ *from* **L** *in time* $p(n, m)$*, where* $n$ *is the size of* $\Sigma$ *and* $m$ *is the number of its models (over variables occurring in* $\Sigma$*).*

| Notation | Query |
|---|---|
| **CO** | polytime consistency check |
| **VA** | polytime validity check |
| **CE** | polytime clausal entailment check |
| **IM** | polytime implicant check |
| **EQ** | polytime equivalence check |
| **SE** | polytime sentential entailment check |
| **CT** | polytime model counting |
| **ME** | polytime model enumeration |

Table 4: Notations for queries.

| **L** | **CO** | **VA** | **CE** | **IM** | **EQ** | **SE** | **CT** | **ME** |
|---|---|---|---|---|---|---|---|---|
| NNF | ∘ | ∘ | ∘ | ∘ | ∘ | ∘ | ∘ | ∘ |
| DNNF | √ | ∘ | √ | ∘ | ∘ | ∘ | ∘ | √ |
| d-NNF | ∘ | ∘ | ∘ | ∘ | ∘ | ∘ | ∘ | ∘ |
| s-NNF | ∘ | ∘ | ∘ | ∘ | ∘ | ∘ | ∘ | ∘ |
| f-NNF | ∘ | ∘ | ∘ | ∘ | ∘ | ∘ | ∘ | ∘ |
| d-DNNF | √ | √ | √ | √ | ? | ∘ | √ | √ |
| sd-DNNF | √ | √ | √ | √ | ? | ∘ | √ | √ |
| BDD | ∘ | ∘ | ∘ | ∘ | ∘ | ∘ | ∘ | ∘ |
| FBDD | √ | √ | √ | √ | ? | ∘ | √ | √ |
| OBDD | √ | √ | √ | √ | √ | ∘ | √ | √ |
| $\texttt{OBDD}_<$ | √ | √ | √ | √ | √ | √ | √ | √ |
| DNF | √ | ∘ | √ | ∘ | ∘ | ∘ | ∘ | √ |
| CNF | ∘ | √ | ∘ | √ | ∘ | ∘ | ∘ | ∘ |
| PI | √ | √ | √ | √ | √ | √ | ∘ | √ |
| IP | √ | √ | √ | √ | √ | √ | ∘ | √ |
| MODS | √ | √ | √ | √ | √ | √ | √ | √ |

Table 5: Subsets of the NNF language and their corresponding polytime queries. √ means "satisfies" and ∘ means "does not satisfy unless P = NP."

Table 4 summarizes the queries we are interested in and their acronyms.

The following proposition states what we know about the availability of polytime algorithms for answering the above queries, with respect to all languages we introduced in Section 2.

**Proposition 4.1** *The results in Table 5 hold.*

The results of Proposition 4.1 are summarized in Figure 4. One can draw a number of conclusions based on the results in this figure. First, NNF, s-NNF, d-NNF, f-NNF, and BDD fall in one equivalence class that does not support any polytime queries and CNF satisfies only **VA** and **IM**; hence, none of them qualifies as a target compilation language in this case. But the remaining languages all support polytime tests for consistency and clausal entailment. Therefore, simply imposing either of smoothness (s-NNF), determinism (d-NNF), flatness (f-NNF), or decision (BDD) on the NNF language does not lead to tractability with respect to any of the queries we consider—neither of these properties seem to be significant in isolation. Decomposability (DNNF), however, is an exception and leads immediately to polytime tests for both consistency and clausal entailment, and to a polytime algorithm for model enumeration.

Recall the succinctness ordering DNNF $<$ d-DNNF $<$ FBDD $<$ OBDD $<$ $\text{OBDD}_<$ $<$ MODS from Figure 5. By adding decomposability (DNNF), we obtain polytime tests for consistency and clausal entailment, in addition to a polytime model enumeration algorithm. By adding determinism to decomposability (d-DNNF), we obtain polytime tests for validity, implicant and model counting, which are quite significant. It is not clear, however, whether the combination of decomposability and determinism leads to a polytime test for equivalence. Moreover, adding the decision property on top of decomposability and determinism (FBDD) does not appear to increase tractability with respect to the given queries[3], although it does lead to reducing language succinctness as shown in Figure 5. On the other hand, adding the ordering property on top of decomposability, determinism and decision, leads to polytime tests for equivalence (OBDD and $\text{OBDD}_<$) as well as sentential entailment provided that the ordering $<$ is fixed ($\text{OBDD}_<$).

As for the succinctness ordering NNF $<$ DNNF $<$ DNF $<$ IP $<$ MODS from Figure 5, note that DNNF is obtained by imposing decomposability on NNF, while DNF is obtained by imposing flatness and simple-conjunction (which is stronger than decomposability). What is interesting is that DNF is less succinct than DNNF, yet does not support any more polytime queries; see Figure 4. However, the addition of smoothness (and determinism) on top of flatness and simple-conjunction (MODS) leads to five additional polytime queries, including equivalence and entailment tests.[4]

We close this section by noting that determinism appears to be necessary (but not sufficient) for polytime model counting: only deterministic languages, d-DNNF, sd-DNNF, FBDD, OBDD, $\text{OBDD}_<$ and MODS, support polytime counting. Moreover, polytime counting implies a polytime test of validity, but the opposite is not true.

## 5. Transforming a Compiled Theory

A query is an operation that returns information about a theory without changing it. A transformation, on the other hand, is an operation that returns a modified theory, which is then operated on using queries. Many applications require a combination of transformations and queries.

**Definition 5.1** **($\wedge$C, $\vee$C)** *Let* **L** *be a subset of* NNF. **L** *satisfies* $\wedge$**C** *(*$\vee$**C***) iff there exists a polytime algorithm that maps every finite set of formulas* $\Sigma_1, \ldots, \Sigma_n$ *from* **L** *to a formula of* **L** *that is logically equivalent to* $\Sigma_1 \wedge \ldots \wedge \Sigma_n$ *(*$\Sigma_1 \vee \ldots \vee \Sigma_n$*).*

**Definition 5.2** **($\neg$C)** *Let* **L** *be a subset of* NNF. **L** *satisfies* $\neg$**C** *iff there exists a polytime algorithm that maps every formula* $\Sigma$ *from* **L** *to a formula of* **L** *that is logically equivalent to* $\neg\Sigma$.

If a language satisfies one of the above properties, we will say that it is *closed* under the corresponding operator. Closure under logical connectives is important for two key reasons. First, it has implications on how compilers are constructed for a given target language. For example, if a clause can be easily compiled into some language **L**, then closure under conjunction implies that compiling a CNF sentence into **L** is easy. Second, it has implications on the class of polytime queries supported by the target language: If a language **L** satisfies **CO** and is closed under negation and conjunction, then it must satisfy **SE** (to test whether $\Delta \models \Gamma$, all we have to do, by the Refutation Theorem, is test whether $\Delta \wedge \neg\Gamma$ is inconsistent). Similarly, if a language satisfies **VA** and is closed under negation and disjunction, it must satisfy **SE** by the Deduction Theorem.

3. Deciding the equivalence of two sentences in FBDD, d-DNNF, or in sd-DNNF, can be easily shown to be in coNP. However, we do not have a proof of coNP-hardness, nor do we have deterministic polytime algorithms for deciding these problems. Actually, the latter case is quite unlikely as the equivalence problem for FBDD has been intensively studied, with no such algorithm in sight. Note, however, that the equivalence of two sentences in FBDD can be decided probabilistically in polytime (Blum, Chandra, & Wegman, 1980), and similarly for sentences in d-DNNF (Darwiche & Huang, 2002).
4. Given flatness, simple-conjunction and smoothness, we can obtain determinism by simply removing duplicated terms.

It is important to stress here that some languages are closed under a logical operator, only if the number of operands is bounded by a constant. We will refer to this as *bounded closure.*

**Definition 5.3 ($\wedge$BC, $\vee$BC)** *Let* **L** *be a subset of* NNF. **L** *satisfies* $\wedge$**BC** *(*$\vee$**BC***) iff there exists a polytime algorithm that maps every pair of formulas $\Sigma$ and $\Phi$ from* **L** *to a formula of* **L** *that is logically equivalent to $\Sigma \wedge \Phi$ ($\Sigma \vee \Phi$).*

We now turn to another important transformation:

**Definition 5.4 (Conditioning)** *(Darwiche, 1999) Let $\Sigma$ be a propositional formula, and let $\gamma$ be a consistent term. The conditioning of $\Sigma$ on $\gamma$, noted $\Sigma \mid \gamma$, is the formula obtained by replacing each variable $X$ of $\Sigma$ by true (resp. false) if $X$ (resp. $\neg X$) is a positive (resp. negative) literal of $\gamma$.*

**Definition 5.5 (CD)** *Let* **L** *be a subset of* NNF. **L** *satisfies* **CD** *iff there exists a polytime algorithm that maps every formula $\Sigma$ from* **L** *and every consistent term $\gamma$ to a formula from* **L** *that is logically equivalent to $\Sigma \mid \gamma$.*

Conditioning has a number of applications, and corresponds to *restriction* in the literature on Boolean functions. The main application of conditioning is due to a theorem, which says that $\Sigma \wedge \gamma$ is consistent iff $\Sigma \mid \gamma$ is consistent (Darwiche, 2001a, 1999). Therefore, if a language satisfies **CO** and **CD**, then it must also satisfy **CE**. Conditioning also plays a key role in building compilers that enforce decomposability. If two sentences $\Delta_1$ and $\Delta_2$ are both decomposable (belong to DNNF), their conjunction $\Delta_1 \wedge \Delta_2$ is not necessarily decomposable since the sentences may share variables. Conditioning can be used to ensure decomposability in this case since $\Delta_1 \wedge \Delta_2$ is equivalent to $\bigvee_\gamma (\Delta_1 \mid \gamma) \wedge (\Delta_2 \mid \gamma) \wedge \gamma$, where $\gamma$ is a term covering all variables shared by $\Delta_1$ and $\Delta_2$. Note that $\bigvee_\gamma (\Delta_1 \mid \gamma) \wedge (\Delta_2 \mid \gamma) \wedge \gamma$ must be decomposable since $\Delta_1 \mid \gamma$ and $\Delta_2 \mid \gamma$ do not mention variables in $\gamma$. The previous proposition is indeed a generalization to multiple variables of the well-known Shannon expansion in the literature on Boolean functions. It is also the basis for compiling CNF into DNNF (Darwiche, 1999, 2001a).

Another critical transformation we shall consider is that of *forgetting* (also referred to as marginalization, or elimination of middle terms (Boole, 1854)):

**Definition 5.6 (Forgetting)** *Let $\Sigma$ be a propositional formula, and let* **X** *be a subset of variables from PS. The* forgetting *of* **X** *from $\Sigma$, denoted $\exists \mathbf{X}.\Sigma$, is a formula that does not mention any variable from* **X** *and for every formula $\alpha$ that does not mention any variable from* **X***, we have $\Sigma \models \alpha$ precisely when $\exists \mathbf{X}.\Sigma \models \alpha$.*

Therefore, to forget variables from **X** is to remove any reference to **X** from $\Sigma$, while maintaining all information that $\Sigma$ captures about the complement of **X**. Note that $\exists \mathbf{X}.\Sigma$ is unique up to logical equivalence.

**Definition 5.7 (FO, SFO)** *Let* **L** *be a subset of* NNF. **L** *satisfies* **FO** *iff there exists a polytime algorithm that maps every formula $\Sigma$ from* **L** *and every subset* **X** *of variables from PS to a formula from* **L** *equivalent to $\exists \mathbf{X}.\Sigma$. If the property holds for singleton* **X***, we say that* **L** *satisfies* **SFO***.*

Forgetting is an important transformation as it allows us to focus/project a theory on a set of variables. For example, if we know that some variables **X** will never appear in entailment queries, we can forget these variables from the compiled theory while maintaining its ability to answer such queries correctly. Another application of forgetting is in counting/enumerating the instantiations of some variables **Y**, which are consistent with a theory $\Delta$. This query can be answered by counting/enumerating the models of $\exists \mathbf{X}.\Delta$, where **X** is the complement of **Y**. Forgetting also has applications to planning, diagnosis and belief revision. For instance, in the SATPLAN framework, compiling away fluents or actions amounts to forgetting variables. In model-based diagnosis, compiling away every variable except the abnormality ones does not remove any piece of information required to compute the conflicts and the diagnoses of a system (Darwiche, 2001a). Forgetting has also been used to design update operators with valuable properties (Herzig & Rifi, 1999).

| Notation | Transformation |
|---|---|
| **CD** | polytime conditioning |
| **FO** | polytime forgetting |
| **SFO** | polytime singleton forgetting |
| $\wedge$**C** | polytime conjunction |
| $\wedge$**BC** | polytime bounded conjunction |
| $\vee$**C** | polytime disjunction |
| $\vee$**BC** | polytime bounded disjunction |
| $\neg$**C** | polytime negation |

Table 6: Notations for transformations.

| **L** | **CD** | **FO** | **SFO** | $\wedge$**C** | $\wedge$**BC** | $\vee$**C** | $\vee$**BC** | $\neg$**C** |
|---|---|---|---|---|---|---|---|---|
| NNF | √ | ◦ | √ | √ | √ | √ | √ | √ |
| DNNF | √ | √ | √ | ◦ | ◦ | √ | √ | ◦ |
| d-NNF | √ | ◦ | √ | √ | √ | √ | √ | √ |
| s-NNF | √ | ◦ | √ | √ | √ | √ | √ | √ |
| f-NNF | √ | ◦ | √ | • | • | • | • | √ |
| d-DNNF | √ | ◦ | ◦ | ◦ | ◦ | ◦ | ◦ | ? |
| sd-DNNF | √ | ◦ | ◦ | ◦ | ◦ | ◦ | ◦ | ? |
| BDD | √ | ◦ | √ | √ | √ | √ | √ | √ |
| FBDD | √ | • | ◦ | • | ◦ | • | ◦ | √ |
| OBDD | √ | • | √ | • | ◦ | • | ◦ | √ |
| $\texttt{OBDD}_<$ | √ | • | √ | • | √ | • | √ | √ |
| DNF | √ | √ | √ | • | √ | √ | √ | • |
| CNF | √ | ◦ | √ | √ | √ | • | √ | • |
| PI | √ | √ | √ | • | • | • | √ | • |
| IP | √ | • | • | • | √ | • | • | • |
| MODS | √ | √ | √ | • | √ | • | • | • |

Table 7: Subsets of the NNF language and their polytime transformations. √ means "satisfies," • means "does not satisfy," while ◦ means "does not satisfy unless P=NP."

Table 6 summarizes the transformations we are interested in and their acronyms. The following proposition states what we know about the tractability of these transformations with respect to the identified target compilation languages.

**Proposition 5.1** *The results in Table 7 hold.*

One can draw a number of observations regarding Table 7. First, all languages we consider satisfy **CD** and, hence, lend themselves to efficient application of the conditioning transformation. As for forgetting multiple variables, only DNNF, DNF, PI and MODS permit that in polytime. It is important to stress here that none of FBDD, OBDD and $\texttt{OBDD}_<$ permits polytime forgetting of multiple variables. This is noticeable since some of the recent applications of $\texttt{OBDD}_<$ to planning—within the so-called symbolic model checking approach to planning (A. Cimmati & Traverso, 1997)—depend crucially on the operation of forgetting and it may be more suitable to use a language that satisfies **FO** in this case. Note, however, that OBDD and $\text{OBDD}_<$ allow the forgetting of a single variable in polytime, but FBDD does not allow even that. d-DNNF is similar to FBDD as it satisfies neither **FO** nor **SFO**.

It is also interesting to observe that none of the target compilation languages is closed under conjunction. A number of them, however, are closed under bounded conjunction, including $\text{OBDD}_<$, DNF, IP and MODS.

As for disjunction, the only target compilation languages that are closed under disjunction are DNNF and DNF. The $\text{OBDD}_<$ and PI languages, however, are closed under bounded disjunction. Again, the d-DNNF, FBDD and OBDD languages are closed under neither.

The only target compilation languages that are closed under negation are FBDD, OBDD and $\text{OBDD}_<$, while it is not known whether d-DNNF or sd-DNNF are closed under this operation. Note that d-DNNF and FBDD support the same set of polytime queries (equivalence checking is unknown for both) so they are indistinguishable from that viewpoint. Moreover, the only difference between the two languages in Table 7 is the closure of FBDD under negation, which does not seem to be that significant in light of no closure under either conjunction or disjunction. Note, however, that d-DNNF is more succinct than FBDD as given in Figure 5.

Finally, $\text{OBDD}_<$ is the only target compilation language that is closed under negation, bounded conjunction, and bounded disjunction. This closure actually plays an important role in compiling propositional theories into $\text{OBDD}_<$ and is the basis of state-of-the-art compilers for this purpose (Bryant, 1986).

## 6. Conclusion

The main contribution of this paper is a methodology for analyzing propositional compilation approaches according to two key dimensions: the succinctness of the target compilation language, and the class of queries and transformations it supports in polytime. The second main contribution of the paper is a comprehensive analysis, according to the proposed methodology, of more than a dozen languages for which we have produced a knowledge compilation map, which cross-ranks these languages according to their succinctness, and the polytime queries and transformations they support. This map allows system designers to make informed decisions on which target compilation language to use: after the class of queries/transformations have been decided based on the application of interest, the designer chooses the most succinct target compilation language that supports such operations in polytime. Another key contribution of this paper is the uniform treatment we have applied to diverse target compilation languages, showing how they all are subsets of the NNF language. Specifically, we have identified a number of simple, yet meaningful, properties, including decomposability, determinism, decision and flatness, and showed how combinations of these properties give rise to different target compilation languages. The studied subsets include some well known languages such as PI, which has been influential in AI; $\text{OBDD}_<$, which has been influential in formal verification; and CNF and DNF, which have been quite influential in computer science. The subsets also include some relatively new languages such as DNNF and d-DNNF, which appear to represent interesting, new balances between language succinctness and query/transformation tractability.

## Acknowledgments

This is a revised and extended version of the paper "A Perspective on Knowledge Compilation," in Proceedings of the $17^{th}$ International Joint Conference on Artificial Intelligence (IJCAI'01), pp. 175-182, 2001. We wish to thank Alvaro del Val, Mark Hopkins, Jérôme Lang and the anonymous reviewers for some suggestions and comments, as well as Ingo Wegener for his help with some of the issues discussed in the paper. This work has been done while the second author was a visiting researcher with the Computer Science Department at UCLA. The first author has been partly supported by NSF grant IIS-9988543 and MURI grant N00014-00-1-0617. The second author has been partly supported by the IUT de Lens, the Université d'Artois, the Nord/Pas-de-Calais Région under the TACT-TIC project, and by the European Community FEDER Program.

## Appendix A. Proofs

To simplify the proofs of our main propositions later on, we have identified a number of lemmas that we list below. Some of the proofs of these lemmas are direct, but we include them for completeness.

**Lemma A.1** *Every sentence in* `d-DNNF` *can be translated to an equivalent sentence in* `sd-DNNF` *in polytime.*

**Proof:** Let $\alpha = \alpha_1 \vee \ldots \vee \alpha_n$ be an or-node in a `d-DNNF` sentence $\Sigma$. Suppose that $\alpha$ is not smooth and let $V = \mathit{Vars}(\alpha)$. Consider now the sentence $\Sigma_s$ obtained by replacing in $\Sigma$ each such node by $\bigvee_{i=1}^{n} \alpha_i \wedge \bigwedge_{v \in V \setminus \mathit{Vars}(\alpha_i)} (\neg v \vee v)$. Then $\Sigma_s$ is equivalent to $\Sigma$ and is smooth. Moreover, $\Sigma_s$ can be computed in time polynomial in the size of $\Sigma$ and it satisfies decomposability and determinism. □

**Lemma A.2** *Every sentence in* `FBDD` *can be translated to an equivalent sentence in* `FBDD` $\cap$ `s-NNF` *in polytime.*

**Proof:** Let $\Sigma$ be a sentence in `FBDD` and let $\alpha$ be a node in $\Sigma$. We can always replace $\alpha$ with $(Y \wedge \alpha) \vee (\neg Y \wedge \alpha)$, for some variable $Y$, while preserving equivalence and the decision property. Moreover, as long as the variable $Y$ does not appear in $\alpha$ and is not an ancestor of $\alpha$, then decomposability is also preserved (that is, the resulting sentence is in `FBDD`). Note here that "ancestor" is with respect to the binary decision diagram notation of $\Sigma$–see left of Figure 2.

Now, suppose that $(X \wedge \alpha) \vee (\neg X \wedge \beta)$ is an or-node in $\Sigma$. Suppose further that the or-node is not smooth. Hence, there is some $Y$ which appears in $\mathit{Vars}(\beta)$ but not in $\mathit{Vars}(\alpha)$ (or the other way around). Since $\Sigma$ is decomposable, then $Y$ cannot be an ancestor of $\alpha$ (since in that case it would also be an ancestor of $\beta$, which is impossible by decomposability of $\Sigma$). Hence, we can replace $\alpha$ with $(Y \wedge \alpha) \vee (\neg Y \wedge \alpha)$, while preserving equivalence, decision and decomposability. By repeating the above process, we can smooth $\Sigma$ while preserving all the necessary properties. Finally, note that for every or-node $(X \wedge \alpha) \vee (\neg X \wedge \beta)$ in $\Sigma$, we need to repeat the above process at most $\mid \mathit{Vars}(\alpha) - \mathit{Vars}(\beta) \mid + \mid \mathit{Vars}(\beta) - \mathit{Vars}(\alpha) \mid$ times. Hence, the smoothing operation can be performed in polytime. □

**Lemma A.3** *If a subset* **L** *of* `NNF` *satisfies* **CO** *and* **CD***, then it also satisfies* **ME**.

**Proof:** Let $\Sigma$ be a sentence in **L**. First, we test if $\Sigma$ is inconsistent (can be done in polytime). If it is, we return the empty set of models. Otherwise, we construct a decision-tree representation of the models of $\Sigma$. Given an ordering of the variables $x_1, \ldots, x_n$ of $\mathit{Vars}(\Sigma)$, we start with a tree $T$ consisting of a single root node. For $i = 1$ to $n$, we repeat the following for each leaf node $\alpha$ (corresponds to a consistent term) in $T$:

a. If $\Sigma \mid \alpha \wedge x_i$ is consistent, we add $x_i$ as a child to $\alpha$;

b. If $\Sigma \mid \alpha \wedge \neg x_i$ is consistent, we add $\neg x_i$ as a child to $\alpha$.

The key points are:

- Test (a) and Test (b) can be performed in time polynomial in the size of $\Sigma$ (since **L** satisfies **CO** and **CD**).

- Either Test (a) or Test (b) above must succeed (since $\Sigma$ is consistent).

Hence, the number of tests performed is $\mathcal{O}(mn)$, where $m$ is the number of leaf nodes in the final decision tree (bounded by the number of models of $\Sigma$) and $n$ is the number of variables of $\Sigma$. □

**Lemma A.4** *If a subset of* `NNF` *satisfies* **CO** *and* **CD**, *then it also satisfies* **CE**.

**Proof:** To test whether sentence $\Sigma$ entails non-valid clause $\alpha$, $\Sigma \models \alpha$, it suffices to test whether $\Sigma \mid \neg\alpha$ is inconsistent (Darwiche, 2001a). □

**Lemma A.5** *Let $\alpha$ and $\beta$ be two sentences that share no variables. Then $\alpha \vee \beta$ is valid iff $\alpha$ is valid or $\beta$ is valid.*

**Proof:** $\alpha \vee \beta$ is valid iff $\neg\alpha \wedge \neg\beta$ is inconsistent. Since $\neg\alpha$ and $\neg\beta$ share no variables, then $\neg\alpha \wedge \neg\beta$ is inconsistent iff $\neg\alpha$ is inconsistent or $\neg\beta$ is. This is true iff $\alpha$ is valid or $\beta$ is valid. □

**Lemma A.6** *Let $\Sigma$ be a sentence in* `d-DNNF` *and let $\gamma$ be a clause. Then a sentence in* `d-DNNF` *which is equivalent to $\Sigma \vee \gamma$ can be constructed in polytime in the size of $\Sigma$ and $\gamma$.*

**Proof:** Let $l_1, \ldots, l_n$ be the literals that appear in clause $\gamma$. Then $\beta = \bigvee_{i=1}^{n}(l_i \wedge \bigwedge_{j=1}^{i-1} \neg l_j)$ is equivalent to clause $\gamma$, is in `d-DNNF`, and can be constructed in polytime in size of $\gamma$. Now let $\alpha$ be the term equivalent to $\neg\gamma$. We have that $\Sigma \vee \gamma$ is equivalent to $((\Sigma \mid \alpha) \wedge \alpha) \vee \beta$. The last sentence is in `d-DNNF` and can be constructed in polytime in size of $\Sigma$ and $\gamma$. □

**Lemma A.7** *If a subset of* `NNF` *satisfies* **VA** *and* **CD**, *then it also satisfies* **IM**.

**Proof:** To test whether a consistent term $\alpha$ entails sentence $\Sigma$, $\alpha \models \Sigma$, it suffices to test whether $\neg\alpha \vee \Sigma$ is valid. This sentence is equivalent to $\neg\alpha \vee (\alpha \wedge \Sigma)$, to $\neg\alpha \vee (\alpha \wedge (\Sigma \mid \alpha))$, and to $\neg\alpha \vee (\Sigma \mid \alpha)$. Since $\neg\alpha$ and $\Sigma \mid \alpha$ share no variables, the disjunction is valid iff $\neg\alpha$ is valid or $\Sigma \mid \alpha$ is valid (by Lemma A.5). $\neg\alpha$ cannot be valid since $\alpha$ is consistent. $\Sigma \mid \alpha$ can be constructed in polytime since the language satisfies **CD** and its validity can be tested in polytime since the language satisfies **VA**. □

**Lemma A.8** *Every CNF or DNF formula can be translated to an equivalent sentence in* `BDD` *in polytime.*

**Proof:** It is straightforward to convert a clause or term into an equivalent sentence in `BDD`. In order to generate a `BDD` sentence corresponding to the conjunction (resp. disjunction) of `BDD` sentences $\alpha$ and $\beta$, it is sufficient to replace the 1-sink (resp. 0-sink) of $\alpha$ with the root of $\beta$. □

**Lemma A.9** *If a subset of* `NNF` *satisfies* **EQ**, *then it satisfies* **CO** *and* **VA**.

**Proof:** *true* and *false* belong to every `NNF` subset. $\Sigma$ is inconsistent iff it is equivalent to *false*. $\Sigma$ is valid iff it is equivalent to *true*. □

**Lemma A.10** *If a subset of* `NNF` *satisfies* **SE**, *then it satisfies* **EQ**, **CO** *and* **VA**.

**Proof:** Sentences $\Sigma_1$ and $\Sigma_2$ are equivalent iff $\Sigma_1 \models \Sigma_2$ and $\Sigma_2 \models \Sigma_1$. **EQ** implies **CO** and **VA** (Lemma A.9). □

**Lemma A.11** *Let $\Sigma$ be a sentence in* d-DNNF *and let $\gamma$ be a clause. The validity of $\Sigma \vee \gamma$ can be tested in time polynomial in the size of $\Sigma$ and $\gamma$.*

**Proof:** Construct $\Sigma \vee \gamma$ in polytime as given in Lemma A.6 and check its validity, which can be done in polytime too. □

**Lemma A.12** *For every propositional formula $\Sigma$ and every consistent term $\gamma$, we have $\Sigma|\gamma$ is equivalent to*
*$\exists \mathit{Vars}(\gamma).(\Sigma \wedge \gamma)$.*

**Proof:** Without loss of generality, assume that $\Sigma$ is given by the disjunctively-interpreted set of its models (over $\mathit{Vars}(\Sigma)$). Conditioning $\Sigma$ on $\gamma$ leads (1) to removing every model of $\neg\gamma$, then (2) projecting the remaining models so that every variable of $\gamma$ is removed. Conjoining $\Sigma$ with $\gamma$ leads exactly to (1), while forgetting every variable of $\gamma$ in the resulting formula leads exactly to (2) (Lang, Liberatore, & Marquis, 2000). □

**Lemma A.13** *Each sentence $\Sigma$ in* f-NNF *can be converted into an equivalent sentence $\Sigma*$ in polynomial time, where $\Sigma* \in$* CNF *or $\Sigma* \in$* DNF.

**Proof:** We consider three cases for the sentence $\Sigma$:

1. The root node of $\Sigma$ is an and-node. In this case, $\Sigma$ can be turned into a CNF sentence $\Sigma*$ in polynomial time by simply ensuring that each or-node in $\Sigma$ is a clause (that is, a disjunction of literals that share no variables). Let $C$ be an or-node in $\Sigma$. Since $\Sigma$ is flat and its root is an and-node, $C$ must be a child of the root of $\Sigma$ and the children of $C$ must be leaves. Hence, we can easily ensure that $C$ is a clause as follows:
   - If we have one edge from $C$ to some leaf $X$ and another edge from $C$ to $\neg X$ ($C$ is valid), we replace the edge from the root to $C$ by an edge from the root to *true*.
   - If we have more than one edge from $C$ to the same leaf node $X$, we keep only one of these edges and delete the rest.
2. The root of $\Sigma$ is an or-node. $\Sigma$ can be turned into a DNF sentence $\Sigma*$ in a dual way.[5]
3. The root of $\Sigma$ is a leaf node. $\Sigma$ is already a CNF sentence.

□

**Lemma A.14** *$\alpha$ is a prime implicant (resp. an essential prime implicant) of sentence $\Sigma$ iff $\neg\alpha$ is a prime implicate (resp. an essential prime implicate) of $\neg\Sigma$.* [6]

**Proof:** This is a folklore result, immediate from the definitions. □

**Proof of Proposition 3.1**

The proof of this proposition is broken down into eight steps. In each step, we prove a number of succinctness relationships between different languages, and then apply transitivity of the succinctness relation to infer even more relationships. Associated with each step of the proof is a table in which we mark all relationships that are proved in that step–we don't show these marks in the very first table though.

| L | NNF | DNNF | d-DNNF | FBDD | OBDD | OBDD$_<$ | DNF | CNF | PI | IP | MODS | sd-DNNF |
|---|---|---|---|---|---|---|---|---|---|---|---|---|
| NNF | ≤ | ≤ | ≤ | ≤ | ≤ | ≤ | ≤ | ≤ | ≤ | ≤ | ≤ | ≤ |
| DNNF | | ≤ | ≤ | ≤ | ≤ | ≤ | ≤ | | | ≤ | ≤ | ≤ |
| d-DNNF | | | ≤ | ≤ | ≤ | ≤ | | | | | ≤ | ≤ |
| FBDD | | | | ≤ | ≤ | ≤ | | | | | | |
| OBDD | | | | | ≤ | ≤ | | | | | | |
| OBDD$_<$ | | | | | | ≤ | | | | | | |
| DNF | | | | | | | ≤ | | | ≤ | ≤ | |
| CNF | | | | | | | | ≤ | ≤ | | | |
| PI | | | | | | | | | ≤ | | | |
| IP | | | | | | | | | | ≤ | | |
| MODS | | | | | | | | | | | ≤ | |
| sd-DNNF | | | | | | | | | | | ≤ | ≤ |

Table 8:

| L | NNF | DNNF | d-DNNF | FBDD | OBDD | OBDD$_<$ | DNF | CNF | PI | IP | MODS | sd-DNNF |
|---|---|---|---|---|---|---|---|---|---|---|---|---|
| NNF | ≤ | ≤ | ≤ | ≤ | ≤ | ≤ | ≤ | ≤ | ≤ | ≤ | ≤ | ≤ |
| DNNF | | ≤ | ≤ | ≤ | ≤ | ≤ | ≤ | | | ≤ | ≤ | ≤ |
| d-DNNF | | | ≤ | ≤ | ≤ | ≤ | | | | | ≤ | ≤ |
| FBDD | | | | ≤ | ≤ | ≤ | | | | | | |
| OBDD | | | | | ≤ | ≤ | | | | | | |
| OBDD$_<$ | | | | | | ≤ | | | | | | |
| DNF | **≰** | | | | | | ≤ | **≰** | **≰** | ≤ | ≤ | |
| CNF | **≰** | **≰** | | | | | **≰** | ≤ | ≤ | **≰** | | |
| PI | **≰** | **≰** | | | | | **≰** | **≰** | ≤ | **≰** | | |
| IP | **≰** | **≰** | | | | | **≰** | **≰** | **≰** | ≤ | | |
| MODS | | | | | | | | | | | ≤ | |
| sd-DNNF | | | | | | | | | | | ≤ | ≤ |

Table 9:

**Table 8:** Follows immediately from the language inclusions reported in Figure 4.

**Table 9:** We can prove both that DNF $\not\leq$ PI and CNF $\not\leq$ IP (this slightly generalizes the results DNF $\not\leq$ CNF and CNF $\not\leq$ DNF given in (Gogic et al., 1995)).

Let us consider the CNF formula $\Sigma_n = \bigwedge_{i=0}^{n-1}(x_{2i} \vee x_{2i+1})$. This formula is in prime implicates form[7] (and each clause in $\Sigma_n$ is an essential prime implicate of it). Hence its negation $\neg\Sigma_n$ is in prime implicants form (as an easy consequence of Lemma A.14).

Since Quine's early work (Quine, 1959), we know that the number of essential prime implicants (resp. prime implicates) of a formula is a lower bound of the number of terms (resp. clauses) that can be found in any DNF (resp. CNF) representation of it (indeed, any such representation must include the essential prime). $\Sigma_n$ has $2^n$ essential prime implicants. Indeed, this can be easily shown by induction on $n$ given that (i) every literal occurring in $\Sigma_n$ occurs only once, (ii) the set of prime implicants of any nontautological clause is the set of literals occurring in it (up to logical equivalence), and (iii) the distribution property for prime implicants (see e.g., (dual of) Proposition 40 in (Marquis, 2000)) which states that $IP(\alpha \wedge \beta) = max(\{\texttt{PI}_\alpha \wedge \texttt{PI}_\beta \mid \texttt{PI}_\alpha \in IP(\alpha), \texttt{PI}_\beta \in IP(\beta)\}, \models)$ (up to logical equivalence). Subsequently, $\neg\Sigma_n$ has $2^n$ essential prime implicates (cf. Lemma A.14). Accordingly, we obtain that both DNF $\not\leq$ PI and CNF $\not\leq$ IP. We also obtain PI $\not\leq$ IP and IP $\not\leq$ PI. Now, it is well–known that some DNF formulas have exponentially many prime implicants (see the proof of Proposition 5.1 where we show that IP does not satisfy **SFO**). Hence, their negations are CNF

5. Note that f-NNF satisfies $\neg$**C** and that the negation of a CNF sentence (resp. DNF sentence) can be turned into a DNF (resp. CNF) in linear time.
6. A prime implicant (resp. a prime implicate) $\alpha$ of $\Sigma$ is *essential* iff the disjunction (resp. conjunction) of all prime implicants (resp. prime implicates) of $\Sigma$ except $\alpha$ is not equivalent to $\Sigma$.
7. The correctness of (the dual of) Quine's consensus algorithm for computing prime implicants (Quine, 1955) ensures it, since no clause of $\Sigma_n$ is subsumed by another clause and no consensi can be performed since there are no negated variables.

| **L** | NNF | DNNF | d-DNNF | FBDD | OBDD | $\mathtt{OBDD}_<$ | DNF | CNF | PI | IP | MODS | sd-DNNF |
|---|---|---|---|---|---|---|---|---|---|---|---|---|
| NNF | $\le$ | $\le$ | $\le$ | $\le$ | $\le$ | $\le$ | $\le$ | $\le$ | $\le$ | $\le$ | $\le$ | $\le$ |
| DNNF | | $\le$ | $\le$ | $\le$ | $\le$ | $\le$ | $\le$ | | | $\le$ | $\le$ | $\le$ |
| d-DNNF | | | $\le$ | $\le$ | $\le$ | $\le$ | | | | | $\le$ | $\le$ |
| FBDD | | | | $\le$ | $\le$ | $\le$ | | | | | | |
| OBDD | | | | | $\le$ | $\le$ | | | | | | |
| $\mathtt{OBDD}_<$ | | | | | | $\le$ | | | | | | |
| DNF | $\not\le$ | **$\not\le$** | **$\not\le$** | **$\not\le$** | **$\not\le$** | **$\not\le$** | $\le$ | $\not\le$ | $\not\le$ | $\le$ | $\le$ | |
| CNF | $\not\le$ | $\not\le$ | **$\not\le$** | **$\not\le$** | **$\not\le$** | **$\not\le$** | $\not\le$ | $\le$ | $\le$ | $\not\le$ | | |
| PI | $\not\le$ | $\not\le$ | **$\not\le$** | **$\not\le$** | **$\not\le$** | **$\not\le$** | $\not\le$ | $\not\le$ | $\le$ | $\not\le$ | | |
| IP | $\not\le$ | $\not\le$ | **$\not\le$** | **$\not\le$** | **$\not\le$** | **$\not\le$** | $\not\le$ | $\not\le$ | $\not\le$ | $\le$ | | |
| MODS | | | | | | | | | | | $\le$ | |
| sd-DNNF | | | | | | | | | | | $\le$ | $\le$ |

Table 10:

| **L** | NNF | DNNF | d-DNNF | FBDD | OBDD | $\mathtt{OBDD}_<$ | DNF | CNF | PI | IP | MODS | sd-DNNF |
|---|---|---|---|---|---|---|---|---|---|---|---|---|
| NNF | $\le$ | $\le$ | $\le$ | $\le$ | $\le$ | $\le$ | $\le$ | $\le$ | $\le$ | $\le$ | $\le$ | $\le$ |
| DNNF | | $\le$ | $\le$ | $\le$ | $\le$ | $\le$ | $\le$ | | | $\le$ | $\le$ | $\le$ |
| d-DNNF | | | $\le$ | $\le$ | $\le$ | $\le$ | | | | | $\le$ | $\le$ |
| FBDD | **$\not\le$** | **$\not\le$** | **$\not\le$** | $\le$ | $\le$ | $\le$ | | | | | | |
| OBDD | **$\not\le$** | **$\not\le$** | **$\not\le$** | **$\not\le$** | $\le$ | $\le$ | | | | | | |
| $\mathtt{OBDD}_<$ | **$\not\le$** | **$\not\le$** | **$\not\le$** | **$\not\le$** | **$\not\le$** | $\le$ | | | | | | |
| DNF | $\not\le$ | $\not\le$ | $\not\le$ | $\not\le$ | $\not\le$ | $\not\le$ | $\le$ | $\not\le$ | $\not\le$ | $\le$ | $\le$ | |
| CNF | $\not\le$ | $\not\le$ | $\not\le$ | $\not\le$ | $\not\le$ | $\not\le$ | $\not\le$ | $\le$ | $\le$ | $\not\le$ | | |
| PI | $\not\le$ | $\not\le$ | $\not\le$ | $\not\le$ | $\not\le$ | $\not\le$ | $\not\le$ | $\not\le$ | $\le$ | $\not\le$ | | |
| IP | $\not\le$ | $\not\le$ | $\not\le$ | $\not\le$ | $\not\le$ | $\not\le$ | $\not\le$ | $\not\le$ | $\not\le$ | $\le$ | | |
| MODS | | | | | | | | | | | $\le$ | |
| sd-DNNF | | | | | | | | | | | $\le$ | $\le$ |

Table 11:

formulas having exponentially many prime implicates. Subsequently IP $\not\le$ DNF and PI $\not\le$ CNF. The remaining results in this table follow from the transitivity of $\le$.

**Table 10:** The parity function $O_n = \bigoplus_{i=0}^{n-1} x_i$ has linear size $\text{OBDD}_<$ representations (Bryant, 1986) but only exponential size CNF and DNF representations. The reason is that $O_n$ has $2^n$ essential prime implicants (resp. essential prime implicates) and the number of essential prime implicants (resp. essential prime implicates) of a formula is a lower bound of the size of any of its DNF (resp. CNF) representation. This easily shows that both CNF $\not\le$ OBDD and DNF $\not\le$ OBDD. The remaining results in this table follow from the language inclusions reported in Figure 4.

**Table 11:** It is shown in (Darwiche, 2001b) that there is a sentence in d-DNNF which only has exponential FBDD representations. Accordingly, we have FBDD $\not\le$ d-DNNF. In (Gergov & Meinel, 1994a), it is shown that OBDD $\not\le$ FBDD. Finally, it is easy to show that $\mathtt{OBDD}_<$ $\not\le$ OBDD (for instance, the formula $\Sigma_n = \bigwedge_{i=1}^{n}(x_i \Leftrightarrow y_i)$ has an $\mathtt{OBDD}_<$ representation of size polynomial in $n$ whenever $<$ satisfies $x_1 < y_1 < x_2 < \ldots < x_n < y_n$, while it has an $\mathtt{OBDD}_<$ representation of size exponential in $n$ provided that $<$ is s.t. $x_1 < x_2 < \ldots < x_n < y_1 < y_2 < \ldots < y_n$). The remaining results in this table follow from the language inclusions reported in Figure 4.

**Table 12:** **L**' $\not\le^*$ **L** means that **L**' $\not\le$ **L** unless the polynomial hierarchy PH collapses. The results in this table follow since the existence of polysize knowledge compilation functions for clausal entailment implies the collapse of the polynomial hierarchy PH (Selman & Kautz, 1996; Cadoli & Donini, 1997). Now, if DNNF $\le$ CNF, then for each sentence $\Sigma$ in CNF there exists a polysize equivalent sentence $\Gamma$ in DNNF. Therefore, we can test whether a clause is entailed by $\Sigma$ in polytime by testing whether the clause is entailed by $\Gamma$. This proves the existence of polysize knowledge compilation functions for clausal entailment, leading to the collapse of the polynomial hierarchy PH. The same is true for d-DNNF and sd-DNNF since all these languages support a polytime clausal entailment test (see Proposition 4.1).

**Table 13:** In (Wegener, 1987) (Theorem 6.2 pp. 436), a family of $n^2$-variable boolean functions $\Sigma$ is pointed out. Provided that every interpretation $I$ over these $n^2$ variables represents a $n$-vertices digraph (for every $1 \le i, j \le n$, we have $I(x_{i,j}) = 1$ iff $(i, j)$ is an arc of the digraph), $\Sigma(I) = 1$ iff the

| L | NNF | DNNF | d-DNNF | FBDD | OBDD | $OBDD_<$ | DNF | CNF | PI | IP | MODS | sd-DNNF |
|---|---|---|---|---|---|---|---|---|---|---|---|---|
| NNF | ≤ | ≤ | ≤ | ≤ | ≤ | ≤ | ≤ | ≤ | ≤ | ≤ | ≤ | ≤ |
| DNNF | ▮ ≰* ▮ | ≤ | ≤ | ≤ | ≤ | ≤ | ≤ | ▮ ≰* ▮ | | ≤ | ≤ | ≤ |
| d-DNNF | ▮ ≰* ▮ | | ≤ | ≤ | ≤ | ≤ | | ▮ ≰* ▮ | | | ≤ | ≤ |
| FBDD | ≰ | ≰ | ≰ | ≤ | ≤ | ≤ | | | | | | |
| OBDD | ≰ | ≰ | ≰ | ≰ | ≤ | ≤ | | | | | | |
| $OBDD_<$ | ≰ | ≰ | ≰ | ≰ | ≰ | ≤ | | | | | | |
| DNF | ≰ | ≰ | ≰ | ≰ | ≰ | ≰ | ≤ | ≰ | ≰ | ≤ | ≤ | |
| CNF | ≰ | ≰ | ≰ | ≰ | ≰ | ≰ | ≰ | ≤ | ≤ | ≰ | | |
| PI | ≰ | ≰ | ≰ | ≰ | ≰ | ≰ | ≰ | ≰ | ≤ | ≰ | | |
| IP | ≰ | ≰ | ≰ | ≰ | ≰ | ≰ | ≰ | ≰ | ≰ | ≤ | | |
| MODS | | | | | | | | | | | ≤ | |
| sd-DNNF | ▮ ≰* ▮ | | | | | | | ▮ ≰* ▮ | | | ≤ | ≤ |

Table 12:

| L | NNF | DNNF | d-DNNF | FBDD | OBDD | $OBDD_<$ | DNF | CNF | PI | IP | MODS | sd-DNNF |
|---|---|---|---|---|---|---|---|---|---|---|---|---|
| NNF | ≤ | ≤ | ≤ | ≤ | ≤ | ≤ | ≤ | ≤ | ≤ | ≤ | ≤ | ≤ |
| DNNF | ≰* | ≤ | ≤ | ≤ | ≤ | ≤ | ≤ | ≰* | | ≤ | ≤ | ≤ |
| d-DNNF | ≰* | | ≤ | ≤ | ≤ | ≤ | | ≰* | | | ≤ | ≤ |
| FBDD | ≰ | ≰ | ≰ | ≤ | ≤ | ≤ | ▮ ≰ ▮ | ▮ ≰ ▮ | ▮ ≰ ▮ | ▮ ≰ ▮ | | |
| OBDD | ≰ | ≰ | ≰ | ≰ | ≤ | ≤ | ▮ ≰ ▮ | ▮ ≰ ▮ | ▮ ≰ ▮ | ▮ ≰ ▮ | | |
| $OBDD_<$ | ≰ | ≰ | ≰ | ≰ | ≰ | ≤ | ▮ ≰ ▮ | ▮ ≰ ▮ | ▮ ≰ ▮ | ▮ ≰ ▮ | | |
| DNF | ≰ | ≰ | ≰ | ≰ | ≰ | ≰ | ≤ | ≰ | ≰ | ≤ | ≤ | |
| CNF | ≰ | ≰ | ≰ | ≰ | ≰ | ≰ | ≰ | ≤ | ≤ | ≰ | | |
| PI | ≰ | ≰ | ≰ | ≰ | ≰ | ≰ | ≰ | ≰ | ≤ | ≰ | | |
| IP | ≰ | ≰ | ≰ | ≰ | ≰ | ≰ | ≰ | ≰ | ≰ | ≤ | | |
| MODS | | | | | | | | | | | ≤ | |
| sd-DNNF | ≰* | | | | | | | ≰* | | | ≤ | ≤ |

Table 13:

digraph represented by $I$ contains a $k$-clique of a special kind ($k$ is a parameter of the family). It is shown that for certain values of $k$ (depending on $n$), every FBDD representation of $\Sigma$ has exponential size. Moreover, it is shown that $\Sigma$ has only a cubic number of prime implicants. This shows that FBDD $\not\leq$ IP, hence FBDD $\not\leq$ DNF. Because FBDD satisfies $\neg$**C** (see Proposition 5.1),[8] it cannot be the case that $\neg\Sigma$ has a polynomial size FBDD. Since $\neg\Sigma$ has only a cubic number of prime implicates, we obtain that FBDD $\not\leq$ PI, hence FBDD $\not\leq$ CNF. The remaining results in this table follow since FBDD$\leq$ OBDD $\leq$ $OBDD_<$.

**Table 14:** Assume that d-DNNF $\leq$ DNF holds. As a consequence, every sentence $\Sigma$ in DNF can be compiled into an equivalent d-DNNF sentence $\Sigma*$ of polynomial size. Now, checking whether a clause $\gamma$ is entailed by the CNF sentence $\Sigma$ is equivalent to checking whether the DNF sentence $\neg\Sigma \vee \gamma$ is valid. Checking whether $(\neg\Sigma)*\vee\gamma$ is valid—when $(\neg\Sigma)*$ is a d-DNNF sentence and $\gamma$ is a clause—can be achieved in polynomial time by Lemma A.11. Therefore, $(\neg\Sigma)*$ is a polysize compilation of the

8. That is, a sentence in FBDD can be negated in polytime to yield a sentence in FBDD too.

| L | NNF | DNNF | d-DNNF | FBDD | OBDD | $OBDD_<$ | DNF | CNF | PI | IP | MODS | sd-DNNF |
|---|---|---|---|---|---|---|---|---|---|---|---|---|
| NNF | ≤ | ≤ | ≤ | ≤ | ≤ | ≤ | ≤ | ≤ | ≤ | ≤ | ≤ | ≤ |
| DNNF | ≰* | ≤ | ≤ | ≤ | ≤ | ≤ | ≤ | ≰* | | ≤ | ≤ | ≤ |
| d-DNNF | ≰* | ▮ ≰* ▮ | ≤ | ≤ | ≤ | ≤ | ▮ ≰* ▮ | ≰* | | | ≤ | ≤ |
| FBDD | ≰ | ≰ | ≰ | ≤ | ≤ | ≤ | ≰ | ≰ | ≰ | ≰ | | ▮ ≰ ▮ |
| OBDD | ≰ | ≰ | ≰ | ≰ | ≤ | ≤ | ≰ | ≰ | ≰ | ≰ | | ▮ ≰ ▮ |
| $OBDD_<$ | ≰ | ≰ | ≰ | ≰ | ≰ | ≤ | ≰ | ≰ | ≰ | ≰ | | ▮ ≰ ▮ |
| DNF | ≰ | ≰ | ≰ | ≰ | ≰ | ≰ | ≤ | ≰ | ≰ | ≤ | ≤ | ▮ ≰ ▮ |
| CNF | ≰ | ≰ | ≰ | ≰ | ≰ | ≰ | ≰ | ≤ | ≤ | ≰ | | ▮ ≰ ▮ |
| PI | ≰ | ≰ | ≰ | ≰ | ≰ | ≰ | ≰ | ≰ | ≤ | ≰ | | ▮ ≰ ▮ |
| IP | ≰ | ≰ | ≰ | ≰ | ≰ | ≰ | ≰ | ≰ | ≰ | ≤ | | ▮ ≰ ▮ |
| MODS | | | | | | | | | | | ≤ | |
| sd-DNNF | ≰* | ▮ ≰* ▮ | ▮ ≤ ▮ | ▮ ≤ ▮ | ▮ ≤ ▮ | ▮ ≤ ▮ | ▮ ≰* ▮ | ≰* | | | ≤ | ≤ |

Table 14:

| L | NNF | DNNF | d-DNNF | FBDD | OBDD | $\text{OBDD}_<$ | DNF | CNF | PI | IP | MODS | sd-DNNF |
|---|---|---|---|---|---|---|---|---|---|---|---|---|
| NNF | $\leq$ | $\leq$ | $\leq$ | $\leq$ | $\leq$ | $\leq$ | $\leq$ | $\leq$ | $\leq$ | $\leq$ | $\leq$ | $\leq$ |
| DNNF | $\not\leq^*$ | $\leq$ | $\leq$ | $\leq$ | $\leq$ | $\leq$ | $\leq$ | $\not\leq^*$ | | $\leq$ | $\leq$ | $\leq$ |
| d-DNNF | $\not\leq^*$ | $\not\leq^*$ | $\leq$ | $\leq$ | $\leq$ | $\leq$ | $\not\leq^*$ | $\not\leq^*$ | | | $\leq$ | $\leq$ |
| FBDD | $\not\leq$ | $\not\leq$ | $\not\leq$ | $\leq$ | $\leq$ | $\leq$ | $\not\leq$ | $\not\leq$ | $\not\leq$ | $\not\leq$ | ▮ $\leq$ ▮ | $\not\leq$ |
| OBDD | $\not\leq$ | $\not\leq$ | $\not\leq$ | $\not\leq$ | $\leq$ | $\leq$ | $\not\leq$ | $\not\leq$ | $\not\leq$ | $\not\leq$ | ▮ $\leq$ ▮ | $\not\leq$ |
| $\text{OBDD}_<$ | $\not\leq$ | $\not\leq$ | $\not\leq$ | $\not\leq$ | $\not\leq$ | $\leq$ | $\not\leq$ | $\not\leq$ | $\not\leq$ | $\not\leq$ | ▮ $\leq$ ▮ | $\not\leq$ |
| DNF | $\not\leq$ | $\not\leq$ | $\not\leq$ | $\not\leq$ | $\not\leq$ | $\not\leq$ | $\leq$ | $\not\leq$ | $\not\leq$ | $\leq$ | $\leq$ | $\not\leq$ |
| CNF | $\not\leq$ | $\not\leq$ | $\not\leq$ | $\not\leq$ | $\not\leq$ | $\not\leq$ | $\not\leq$ | $\leq$ | $\leq$ | $\not\leq$ | ▮ $\leq$ ▮ | $\not\leq$ |
| PI | $\not\leq$ | $\not\leq$ | $\not\leq$ | $\not\leq$ | $\not\leq$ | $\not\leq$ | $\not\leq$ | $\not\leq$ | $\leq$ | $\not\leq$ | | $\not\leq$ |
| IP | $\not\leq$ | $\not\leq$ | $\not\leq$ | $\not\leq$ | $\not\leq$ | $\not\leq$ | $\not\leq$ | $\not\leq$ | $\not\leq$ | $\leq$ | ▮ $\leq$ ▮ | $\not\leq$ |
| MODS | ▮ $\not\leq$ ▮ | ▮ $\not\leq$ ▮ | ▮ $\not\leq$ ▮ | ▮ $\not\leq$ ▮ | ▮ $\not\leq$ ▮ | ▮ $\not\leq$ ▮ | ▮ $\not\leq$ ▮ | ▮ $\not\leq$ ▮ | ▮ $\not\leq$ ▮ | ▮ $\not\leq$ ▮ | $\leq$ | ▮ $\not\leq$ ▮ |
| sd-DNNF | $\not\leq^*$ | $\not\leq^*$ | $\leq$ | $\leq$ | $\leq$ | $\leq$ | $\not\leq^*$ | $\not\leq^*$ | | | $\leq$ | $\leq$ |

Table 15:

CNF sentence $\Sigma$, allowing clausal entailment to be achieved in polynomial time. The existence of such $(\neg\Sigma)*$ for every CNF sentence $\Sigma$ implies the collapse of the polynomial hierarchy (Selman & Kautz, 1996; Cadoli & Donini, 1997). Hence, we obtain that d-DNNF $\not\leq^*$ DNF. As a consequence, we also have d-DNNF $\not\leq^*$ DNNF. Finally, since every d-DNNF sentence can be turned in polynomial time into an equivalent sd-DNNF sentence by Lemma A.1, we have sd-DNNF $\leq$ d-DNNF. Moreover, since d-DNNF $\leq$ sd-DNNF, we obtain sd-DNNF $\not\leq^*$ DNF, sd-DNNF $\not\leq^*$ DNNF, sd-DNNF $\leq$ FBDD, sd-DNNF $\leq$ OBDD, sd-DNNF $\leq$ $\text{OBDD}_<$, FBDD $\not\leq$ sd-DNNF, $\text{OBDD}_<$ $\not\leq$ sd-DNNF, DNF $\not\leq$ sd-DNNF, CNF $\not\leq$ sd-DNNF, PI $\not\leq$ sd-DNNF and IP $\not\leq$ sd-DNNF.

**Table 15:** Let us now show that MODS is not less succinct than PI, IP, sd-DNNF and OBDD. First, let us consider the formula $\Sigma = \bigvee_{i=1}^{n} x_i$. $\Sigma$ can be represented by PI, IP, sd-DNNF and OBDD formulas of size polynomial in $n$. Contrastingly, $\Sigma$ cannot be represented by a MODS formula of size polynomial in $n$ since $\Sigma$ has $2^n - 1$ models over $Vars(\Sigma)$. Now, it is well-known that the old good Quine-McCluskey's algorithm for generating prime implicants from a MODS representation of a propositional formula $\Sigma$ runs in time polynomial in the number of models of $\Sigma$ (Wegener, 1987). This shows that IP $\leq$ MODS. As to CNF and $\text{OBDD}_<$, it is obvious that a decision tree (or Shannon tree) for $\Sigma$ that respects a given total ordering over $Vars(\Sigma)$ can be generated in polynomial time from a MODS representation of $\Sigma$. Such a decision tree has $m$ 1-leaves where $m$ is the number of models of $\Sigma$ over $Vars(\Sigma)$. Accordingly, it has at most $n * m$ 0-leaves where $n = |Vars(\Sigma)|$. Since the set of all paths from the root of the tree to any 0-leaf can be read as a CNF representation of $\Sigma$, we obtain that CNF $\leq$ MODS. On the other hand, since reducing a decision tree to derive a corresponding $\text{OBDD}_<$ can be done in polynomial time, it follows that an $\text{OBDD}_<$ representation of $\Sigma$ can also be generated from a MODS representation of it. Hence, $\text{OBDD}_<$ $\leq$ MODS. The remaining results in this table follow from the language inclusions reported in Figure 4. □

## Proof of Proposition 4.1

The proof of this proposition is broken down into twelve steps. In each step, we prove a number of results. Associated with each step of the proof is a table in which we mark all results that are proved in that step. The table of the last step includes all results declared by this proposition.

**Table 16:** Every classical CNF or DNF formula can be translated in a straightforward way into an equivalent f-NNF sentence (with a tree structure) in polytime. Moreover, every NNF sentence can be translated into an equivalent s-NNF sentence in polytime (Lemma A.1). Given that **CO** is NP-hard (resp. **VA** is coNP-hard) for classical CNF (resp. DNF) sentences, and the inclusion between the various NNF subsets reported in Figure 4, we obtain the table.

**Table 17:** **SE** implies both **CO** and **VA** (Lemma A.10). Moreover, since **CT** implies both **CO** and **VA**, **IM** implies **VA** (valid term), and **CE** implies **CO** (inconsistent clause), we obtain the table.

| L | CO | VA | CE | IM | EQ | CT | SE | ME |
|---|---|---|---|---|---|---|---|---|
| NNF | ▌○▌ | ▌○▌ | | | | | | |
| DNNF | | ▌○▌ | | | | | | |
| d-NNF | | | | | | | | |
| d-DNNF | | | | | | | | |
| BDD | | | | | | | | |
| FBDD | | | | | | | | |
| OBDD | | | | | | | | |
| OBDD$_<$ | | | | | | | | |
| DNF | | ▌○▌ | | | | | | |
| CNF | ▌○▌ | | | | | | | |
| PI | | | | | | | | |
| IP | | | | | | | | |
| MODS | | | | | | | | |
| s-NNF | ▌○▌ | ▌○▌ | | | | | | |
| f-NNF | ▌○▌ | ▌○▌ | | | | | | |
| sd-DNNF | | | | | | | | |

Table 16:

| L | CO | VA | CE | IM | EQ | CT | SE | ME |
|---|---|---|---|---|---|---|---|---|
| NNF | ○ | ○ | ▌○▌ | ▌○▌ | | ▌○▌ | ▌○▌ | |
| DNNF | | ○ | | ▌○▌ | | ▌○▌ | ▌○▌ | |
| d-NNF | | | | | | | | |
| d-DNNF | | | | | | | | |
| BDD | | | | | | | | |
| FBDD | | | | | | | | |
| OBDD | | | | | | | | |
| OBDD$_<$ | | | | | | | | |
| DNF | | ○ | | ▌○▌ | | ▌○▌ | ▌○▌ | |
| CNF | ○ | | ▌○▌ | | | ▌○▌ | ▌○▌ | |
| PI | | | | | | | | |
| IP | | | | | | | | |
| MODS | | | | | | | | |
| s-NNF | ○ | ○ | ▌○▌ | ▌○▌ | | ▌○▌ | ▌○▌ | |
| f-NNF | ○ | ○ | ▌○▌ | ▌○▌ | | ▌○▌ | ▌○▌ | |
| sd-DNNF | | | | | | | | |

Table 17:

| L | CO | VA | CE | IM | EQ | CT | SE | ME |
|---|---|---|---|---|---|---|---|---|
| NNF | ○ | ○ | ○ | ○ | | ○ | ○ | |
| DNNF | | ○ | | ○ | | ○ | ○ | |
| d-NNF | | | | | | | | |
| d-DNNF | | | | | | | | |
| BDD | | | | | | | | |
| FBDD | | | | | | | | |
| OBDD | | | | | | | | |
| OBDD$_<$ | | | | | | | | |
| DNF | | ○ | | ○ | | ○ | ○ | |
| CNF | ○ | | ○ | | | ○ | ○ | |
| PI | | | | | | | | |
| IP | | | | | | | | |
| MODS | ▌√▌ | ▌√▌ | | | | ▌√▌ | | |
| s-NNF | ○ | ○ | ○ | ○ | | ○ | ○ | |
| f-NNF | ○ | ○ | ○ | ○ | | ○ | ○ | |
| sd-DNNF | | | | | | | | |

Table 18:

| L | CO | VA | CE | IM | EQ | CT | SE | ME |
|---|---|---|---|---|---|---|---|---|
| NNF | ○ | ○ | ○ | ○ | | ○ | ○ | |
| DNNF | [√] | ○ | [√] | ○ | | ○ | ○ | |
| d-NNF | | | | | | | | |
| d-DNNF | [√] | | [√] | | | | | |
| BDD | | | | | | | | |
| FBDD | | | | | | | | |
| OBDD | | | | | | | | |
| OBDD$_<$ | | | | | | | | |
| DNF | [√] | ○ | [√] | ○ | | ○ | ○ | |
| CNF | ○ | | ○ | | | ○ | ○ | |
| PI | | | | | | | | |
| IP | [√] | | [√] | | | | | |
| MODS | √ | √ | [√] | | | √ | | |
| s-NNF | ○ | ○ | ○ | ○ | | ○ | ○ | |
| f-NNF | ○ | ○ | ○ | ○ | | ○ | ○ | |
| sd-DNNF | [√] | | [√] | | | | | |

Table 19:

| L | CO | VA | CE | IM | EQ | CT | SE | ME |
|---|---|---|---|---|---|---|---|---|
| NNF | ○ | ○ | ○ | ○ | | ○ | ○ | [○] |
| DNNF | √ | ○ | √ | ○ | | ○ | ○ | [√] |
| d-NNF | | | | | | | | |
| d-DNNF | √ | | √ | | | | | [√] |
| BDD | | | | | | | | |
| FBDD | [√] | [√] | [√] | [√] | | [√] | | [√] |
| OBDD | [√] | [√] | [√] | [√] | [√] | [√] | [○] | [√] |
| OBDD$_<$ | [√] | [√] | [√] | [√] | [√] | [√] | [√] | [√] |
| DNF | √ | ○ | √ | ○ | | ○ | ○ | [√] |
| CNF | ○ | | ○ | | | ○ | ○ | [○] |
| PI | | | | | | | | |
| IP | √ | | √ | | | | | [√] |
| MODS | √ | √ | √ | | | √ | | [√] |
| s-NNF | ○ | ○ | ○ | ○ | | ○ | ○ | [○] |
| f-NNF | ○ | ○ | ○ | ○ | | ○ | ○ | [○] |
| sd-DNNF | √ | | √ | | | | | [√] |

Table 20:

**Table 18:** A sentence $\Sigma$ is consistent (resp. valid) iff it has a model (resp. $2^n$ models, where $n = |Vars(\Sigma)|$). Moreover, the number of models of $\Sigma$ is given by the number of edges outgoing from the or-node in any MODS representation of $\Sigma$. Accordingly, **CO**, **VA** and **CT** can be achieved in polynomial time when $\Sigma$ is given by a MODS formula which gives us the table.

**Table 19:**Because DNNF satisfies **CE** (Darwiche, 2001a), **CE** implies **CO** and MODS $\subseteq$ DNF $\subseteq$ DNNF, IP$\subseteq$ DNF and
sd-DNNF $\subseteq$ d-DNNF $\subseteq$ DNNF, we obtain the table.

**Table 20:** We now use the following results:

**CD** and **CO** imply **CE** (Lemma A.4).

**CD** and **VA** imply **IM** (Lemma A.7).

**CD** and **CO** imply **ME** (Lemma A.3).

All considered NNF subsets satify **CD** (cf. Proposition 5.1).

If an NNF subset does not satisfy **CO** it cannot satisfy **ME**.

It is well-known that FBDD satisfies **CO**, **VA** and **CT**, and that OBDD$_<$ satisfies (in addition) **EQ** (Gergov & Meinel, 1994a; Bryant, 1992).

Since $\Sigma \models \alpha$ holds iff $\Sigma \wedge \neg\alpha$ is inconsistent and since OBDD$_<$ satisfies **CO**, $\neg$**C** and $\wedge$**BC** (cf. Proposition 5.1), OBDD$_<$ also satisfies **SE**.

| L | CO | VA | CE | IM | EQ | CT | SE | ME |
|---|---|---|---|---|---|---|---|---|
| NNF | ∘ | ∘ | ∘ | ∘ | | ∘ | ∘ | ∘ |
| DNNF | √ | ∘ | √ | ∘ | | ∘ | ∘ | √ |
| d-NNF | | | | | | | | |
| d-DNNF | √ | | √ | | | | | √ |
| BDD | | | | | | | | |
| FBDD | √ | √ | √ | √ | | √ | | √ |
| OBDD | √ | √ | √ | √ | √ | √ | ∘ | √ |
| $\texttt{OBDD}_<$ | √ | √ | √ | √ | √ | √ | √ | √ |
| DNF | √ | ∘ | √ | ∘ | | ∘ | ∘ | √ |
| CNF | ∘ | ▌√▌ | ∘ | ▌√▌ | | ∘ | ∘ | ∘ |
| PI | | ▌√▌ | | ▌√▌ | | | | |
| IP | √ | | √ | | | | | √ |
| MODS | √ | √ | √ | | | √ | | √ |
| s-NNF | ∘ | ∘ | ∘ | ∘ | | ∘ | ∘ | ∘ |
| f-NNF | ∘ | ∘ | ∘ | ∘ | | ∘ | ∘ | ∘ |
| sd-DNNF | √ | | √ | | | | | √ |

Table 21:

| L | CO | VA | CE | IM | EQ | CT | SE | ME |
|---|---|---|---|---|---|---|---|---|
| NNF | ∘ | ∘ | ∘ | ∘ | | ∘ | ∘ | ∘ |
| DNNF | √ | ∘ | √ | ∘ | | ∘ | ∘ | √ |
| d-NNF | ▌∘▌ | ▌∘▌ | ▌∘▌ | ▌∘▌ | | ▌∘▌ | ▌∘▌ | ▌∘▌ |
| d-DNNF | √ | | √ | | | | | √ |
| BDD | ▌∘▌ | ▌∘▌ | ▌∘▌ | ▌∘▌ | | ▌∘▌ | ▌∘▌ | ▌∘▌ |
| FBDD | √ | √ | √ | √ | | √ | | √ |
| OBDD | √ | √ | √ | √ | √ | √ | ∘ | √ |
| $\texttt{OBDD}_<$ | √ | √ | √ | √ | √ | √ | √ | √ |
| DNF | √ | ∘ | √ | ∘ | | ∘ | ∘ | √ |
| CNF | ∘ | √ | ∘ | √ | | ∘ | ∘ | ∘ |
| PI | | √ | | √ | | | | |
| IP | √ | | √ | | | | | √ |
| MODS | √ | √ | √ | | | √ | | |
| s-NNF | ∘ | ∘ | ∘ | ∘ | | ∘ | ∘ | ∘ |
| f-NNF | ∘ | ∘ | ∘ | ∘ | | ∘ | ∘ | ∘ |
| sd-DNNF | √ | | √ | | | | | √ |

Table 22:

Obviously enough, any query concerning OBDD is equivalent to the corresponding query concerning $\texttt{OBDD}_<$ provided that only one DAG is brought into play. Together with the above results, we conclude that OBDD satisfies **CO**, **VA** and **CT**. Since this fragment satisfies **CD** as well, it satisfies **CE**, **IM** and **ME** in addition. It also satisfies **EQ** (see Theorem 8.11 from (Meinel & Theobald, 1998)) but does not satisfy **SE** (unless $\mathsf{P} = \mathsf{NP}$). Indeed, it is known that checking the consistency of two $\texttt{OBDD}_<$ formulas $\alpha$ and $\beta$ (based on two different variable orderings $<$) is $\mathsf{NP}$-complete (Lemma 8.14 from (Meinel & Theobald, 1998)). Since OBDD satisfies $\neg$**C** and since $\alpha \land \beta$ is consistent iff $\alpha \not\models \neg\beta$, checking sentential entailment for OBDD formulas is $\mathsf{coNP}$-complete.

These results lead to the table.

**Table 21:** It is known that **IM** is satisfied by classical CNF formulas (hence, PI) (in order to check whether a non-valid clause is implied by a consistent term, it is sufficient to test that they share a literal). CNF (hence, PI) is also known to satisfy **VA**. We then obtain the table.

**Table 22:** Every sentence in CNF or DNF can be turned into an equivalent sentence in BDD in polytime (Lemma A.8). Hence, a $\circ$ in a CNF or DNF cell implies a $\circ$ in the corresponding BDD cell. Similarly, since BDD $\subseteq$ d-NNF, a $\circ$ in a BDD cell implies a $\circ$ in the corresponding d-NNF cell. This leads to the table.

**Table 23:** Since **EQ** implies **CO** and **VA** (Lemma A.9), a $\circ$ in a **CO** or **VA** cell implies a $\circ$ in the corresponding **EQ** cell. This leads to the table.

**Table 24:** By definition, PI satisfies **CE** and IP satisfies **IM**. Since PI $\subseteq$ CNF and IP $\subseteq$ DNF, this implies that both PI and IP satisfy **SE**. Now, **SE** implies **EQ**, hence both PI and IP satisfy **EQ** (actually, two equivalent formulas share the same prime implicates and the same prime implicants (both forms are canonical ones, provided that one representative per equivalence class is considered,

| L | CO | VA | CE | IM | EQ | CT | SE | ME |
|---|---|---|---|---|---|---|---|---|
| NNF | ○ | ○ | ○ | ○ | ▮○▮ | ○ | ○ | ○ |
| DNNF | √ | ○ | √ | ○ | ▮○▮ | ○ | ○ | √ |
| d-NNF | ○ | ○ | ○ | ○ | ▮○▮ | ○ | ○ | ○ |
| d-DNNF | √ | | √ | | | | | √ |
| BDD | ○ | ○ | ○ | ○ | ▮○▮ | ○ | ○ | ○ |
| FBDD | √ | √ | √ | √ | | √ | | √ |
| OBDD | √ | √ | √ | √ | √ | √ | ○ | √ |
| $\text{OBDD}_<$ | √ | √ | √ | √ | √ | √ | √ | √ |
| DNF | √ | ○ | √ | ○ | ▮○▮ | ○ | ○ | √ |
| CNF | ○ | √ | ○ | √ | ▮○▮ | ○ | ○ | ○ |
| PI | | √ | | √ | | | | |
| IP | √ | | √ | | | | | √ |
| MODS | √ | √ | √ | | | √ | | √ |
| s-NNF | ○ | ○ | ○ | ○ | ▮○▮ | ○ | ○ | ○ |
| f-NNF | ○ | ○ | ○ | ○ | ▮○▮ | ○ | ○ | ○ |
| sd-DNNF | √ | | √ | | | | | √ |

Table 23:

| L | CO | VA | CE | IM | EQ | CT | SE | ME |
|---|---|---|---|---|---|---|---|---|
| NNF | ○ | ○ | ○ | ○ | ○ | ○ | ○ | ○ |
| DNNF | √ | ○ | √ | ○ | ○ | ○ | ○ | √ |
| d-NNF | ○ | ○ | ○ | ○ | ○ | ○ | ○ | ○ |
| d-DNNF | √ | | √ | | | | | √ |
| BDD | ○ | ○ | ○ | ○ | ○ | ○ | ○ | ○ |
| FBDD | √ | √ | √ | √ | | √ | | √ |
| OBDD | √ | √ | √ | √ | √ | √ | ○ | √ |
| $\text{OBDD}_<$ | √ | √ | √ | √ | √ | √ | √ | √ |
| DNF | √ | ○ | √ | ○ | ○ | ○ | ○ | √ |
| CNF | ○ | √ | ○ | √ | ○ | ○ | ○ | ○ |
| PI | ▮√▮ | √ | ▮√▮ | √ | ▮√▮ | ▮○▮ | ▮√▮ | ▮√▮ |
| IP | √ | ▮√▮ | √ | ▮√▮ | ▮√▮ | ▮○▮ | ▮√▮ | √ |
| MODS | √ | √ | √ | | | √ | | √ |
| s-NNF | ○ | ○ | ○ | ○ | ○ | ○ | ○ | ○ |
| f-NNF | ○ | ○ | ○ | ○ | ○ | ○ | ○ | ○ |
| sd-DNNF | √ | | √ | | | | | √ |

Table 24:

| L | CO | VA | CE | IM | EQ | CT | SE | ME |
|---|---|---|---|---|---|---|---|---|
| NNF | ○ | ○ | ○ | ○ | ○ | ○ | ○ | ○ |
| DNNF | √ | ○ | √ | ○ | ○ | ○ | ○ | √ |
| d-NNF | ○ | ○ | ○ | ○ | ○ | ○ | ○ | ○ |
| d-DNNF | √ | | √ | | | | | √ |
| BDD | ○ | ○ | ○ | ○ | ○ | ○ | ○ | ○ |
| FBDD | √ | √ | √ | √ | | √ | | √ |
| OBDD | √ | √ | √ | √ | √ | √ | ○ | √ |
| $\text{OBDD}_<$ | √ | √ | √ | √ | √ | √ | √ | √ |
| DNF | √ | ○ | √ | ○ | ○ | ○ | ○ | √ |
| CNF | ○ | √ | ○ | √ | ○ | ○ | ○ | ○ |
| PI | √ | √ | √ | √ | √ | ○ | √ | √ |
| IP | √ | √ | √ | √ | √ | ○ | √ | √ |
| MODS | √ | √ | √ | **√** | **√** | √ | **√** | √ |
| s-NNF | ○ | ○ | ○ | ○ | ○ | ○ | ○ | ○ |
| f-NNF | ○ | ○ | ○ | ○ | ○ | ○ | ○ | ○ |
| sd-DNNF | √ | | √ | | | | | √ |

Table 25:

| L | CO | VA | CE | IM | EQ | CT | SE | ME |
|---|---|---|---|---|---|---|---|---|
| NNF | ○ | ○ | ○ | ○ | ○ | ○ | ○ | ○ |
| DNNF | √ | ○ | √ | ○ | ○ | ○ | ○ | √ |
| d-NNF | ○ | ○ | ○ | ○ | ○ | ○ | ○ | ○ |
| d-DNNF | √ | **√** | √ | **√** | | **√** | | √ |
| BDD | ○ | ○ | ○ | ○ | ○ | ○ | ○ | ○ |
| FBDD | √ | √ | √ | √ | | √ | | √ |
| OBDD | √ | √ | √ | √ | √ | √ | ○ | √ |
| $\text{OBDD}_<$ | √ | √ | √ | √ | √ | √ | √ | √ |
| DNF | √ | ○ | √ | ○ | ○ | ○ | ○ | √ |
| CNF | ○ | √ | ○ | √ | ○ | ○ | ○ | ○ |
| PI | √ | √ | √ | √ | √ | ○ | √ | √ |
| IP | √ | √ | √ | √ | √ | ○ | √ | √ |
| MODS | √ | √ | √ | √ | √ | √ | √ | √ |
| s-NNF | ○ | ○ | ○ | ○ | ○ | ○ | ○ | ○ |
| f-NNF | ○ | ○ | ○ | ○ | ○ | ○ | ○ | ○ |
| sd-DNNF | √ | **√** | √ | **√** | | **√** | | √ |

Table 26:

only)). Since PI satisfies **CE**, it also satisfies **CO**. Since it satisfies **CD** as well (cf. Proposition 5.1), it also satisfies **ME** (Lemma A.3). Contrastingly, the models counting problem for monotone Krom formulas (i.e. conjunctions of clauses containing at most two literals and only positive literals) is #P-complete (Roth, 1996). Such formulas can easily be turned into prime implicates form in polynomial time (Marquis, 2000), hence PI does not satisfy **CT**. Now, since the negation of a formula $\Sigma$ in prime implicates form is a formula in prime implicants form (cf. Lemma A.14), and since the number of models of $\neg\Sigma$ over $Vars(\Sigma)$ is $2^{|Vars(\Sigma)|}$ minus the number of models of $\Sigma$ over $Vars(\Sigma)$, we necessarily have that IP does not satisfy **CT**. This also imply that IP satisfies **VA**, leading to the table.

**Table 25:** In the proof of Proposition 3.1, we have shown that the prime implicants of $\Sigma$ can be computed in polytime from a MODS representation of $\Sigma$. As an immediate consequence, since IP satisfies **IM**, **EQ** and **SE**, we obtain that MODS satisfies **IM**, **EQ** and **SE**, leading to the table.

**Table 26:** Since d-DNNF satisfies **CT** (Darwiche, 2001b), it also satisfies **VA**. Since it satisfies **CD** (Proposition 5.1), it also satisfies **IM** as well (Lemma A.7). Since sd-DNNF $\subseteq$ d-DNNF, these results follow for sd-DNNF. Hence, we obtain the table.

**Table 27:** It is known that determining whether the conjunction of two FBDD formulas $\alpha_1$ and $\alpha_2$ is consistent is NP-complete (Gergov & Meinel, 1994b) Moreover, FBDD satisfies $\neg$**C**. Since $\alpha_1 \wedge \alpha_2$ is inconsistent iff $\alpha_1 \models \neg\alpha_2$, we can reduce the consistency test into an entailment test. Hence, FBDD does not satisfy **SE**. Since FBDD $\subseteq$ d-DNNF, d-DNNF does not satisfy **SE** either. Finally, since every d-DNNF can be translated into an equivalent sd-DNNF sentence in polytime (Lemma A.1), sd-DNNF does not satisfy **SE** either. This leads to the final table above. □

| L | CO | VA | CE | IM | EQ | CT | SE | ME |
|---|---|---|---|---|---|---|---|---|
| NNF | ○ | ○ | ○ | ○ | ○ | ○ | ○ | ○ |
| DNNF | √ | ○ | √ | ○ | ○ | ○ | ○ | √ |
| d-NNF | ○ | ○ | ○ | ○ | ○ | ○ | ○ | ○ |
| d-DNNF | √ | √ | √ | √ | | √ | ▮ ○ ▮ | √ |
| BDD | ○ | ○ | ○ | ○ | ○ | ○ | ○ | ○ |
| FBDD | √ | √ | √ | √ | | √ | ▮ ○ ▮ | √ |
| OBDD | √ | √ | √ | √ | √ | √ | ○ | √ |
| $\texttt{OBDD}_<$ | √ | √ | √ | √ | √ | √ | √ | √ |
| DNF | √ | ○ | √ | ○ | ○ | ○ | ○ | √ |
| CNF | ○ | √ | ○ | √ | ○ | ○ | ○ | ○ |
| PI | √ | √ | √ | √ | √ | ○ | √ | √ |
| IP | √ | √ | √ | √ | √ | ○ | √ | √ |
| MODS | √ | √ | √ | √ | √ | √ | √ | √ |
| s-NNF | ○ | ○ | ○ | ○ | ○ | ○ | ○ | ○ |
| f-NNF | ○ | ○ | ○ | ○ | ○ | ○ | ○ | ○ |
| sd-DNNF | √ | √ | √ | √ | | √ | ▮ ○ ▮ | √ |

Table 27:

### Proof of Proposition 5.1

The proof of this proposition is broken down into eight steps. Each step corresponds to one of the transformations, where we prove all results pertaining to that transformation.

- **CD**. To show that a language **L** satisfies **CD**, we want to show that for any sentence $\Sigma \in \mathbf{L}$ and any consistent term $\gamma$, we can construct in polytime a sentence which belongs to **L** and is equivalent to $\Sigma \mid \gamma$.
  - `NNF`, `f-NNF`, `CNF` and `DNF`. The property is trivially satisfied by these languages: If $\Sigma$ belongs to any of these languages, then replacing the literals of $\gamma$ by a Boolean constant in $\Sigma$ results a sentence in the same language. In the case of `DNF` (resp. `CNF`), some inconsistent terms (valid clauses) may result through conditioning, but these can be removed easily in polynomial time.
  - `DNNF`. It is sufficient to prove that conditioning preserves decomposability. For every propositional sentences $\alpha$, $\beta$ and every consistent term $\gamma$, if $\alpha$ and $\beta$ do not share variables, then $\alpha|\gamma$ and $\beta|\gamma$ do not share variables either since $Vars(\alpha|\gamma) \subseteq Vars(\alpha)$ and $Vars(\beta|\gamma) \subseteq Vars(\beta)$.
  - `d-NNF` and `d-DNNF`. Since `NNF` and `DNNF` satisfy **CD**, it is sufficient to prove that conditioning preserves determinism, i.e. for every propositional formulas $\alpha$, $\beta$ and every consistent term $\gamma$, if $\alpha \wedge \beta \models \mathit{false}$, then $(\alpha|\gamma) \wedge (\beta|\gamma) \models \mathit{false}$. If $\alpha \wedge \beta \models \mathit{false}$, then for every term $\gamma$, we have $(\alpha \wedge \beta) \wedge \gamma \models \mathit{false}$. Since $(\alpha \wedge \beta) \wedge \gamma \equiv ((\alpha \wedge \beta)|\gamma) \wedge \gamma$, this implies that $((\alpha \wedge \beta)|\gamma) \wedge \gamma \models \mathit{false}$. Since $\gamma$ is consistent and share no variable with $(\alpha \wedge \beta)|\gamma$, it must be the case that $(\alpha \wedge \beta)|\gamma$ is inconsistent. This is equivalent to state that $(\alpha|\gamma) \wedge (\beta|\gamma) \models \mathit{false}$.
  - `s-NNF` and `sd-DNNF`. Since `NNF` satisfies **CD**, and since conditioning preserves decomposability and determinism, all we have to show is that conditioning also preserves smoothness. This follows immediately since for two propositional sentences $\alpha$, $\beta$ and a consistent term $\gamma$, we have $Vars(\alpha) = Vars(\beta)$ only if $Vars(\alpha \mid \gamma) = Vars(\beta \mid \gamma)$.
  - `BDD`, `FBDD`, `OBDD` and $\texttt{OBDD}_<$. It is well–known that `BDD` satisfies **CD**—the conditioning operation on binary decision diagrams is known as the *restrict* operation (Bryant, 1986). To condition a sentence $\Sigma$ in `BDD` on a consistent term $\gamma$, we replace every node labeled by a variable in $\gamma$ by one of its two children, according to the sign of the variable in $\gamma$. The resulting sentence is also a `BDD` and is equivalent to $\Sigma \mid \gamma$. The same applies to `FBDD`, `OBDD` and $\texttt{OBDD}_<$.
  - `PI`. The prime implicates of $\Sigma \wedge \gamma$ can be computed in polytime when $\Sigma$ is in prime implicates form and $\gamma$ is a term (see Proposition 36 in (Marquis, 2000)). Moreover, since PI satisfies **FO** (see below), the prime implicates of $\exists\, Vars(\gamma).(\Sigma \wedge \gamma)$ can be computed in polytime. But these are exactly the prime implicates of $\Sigma \mid \gamma$ according to Lemma A.12.

- IP. Let $\Sigma = \bigvee_{i=1}^{n} \gamma_i$ be a formula in prime implicants form. It is clear that the formula $(\bigvee_{i=1}^{n} \gamma_i) \mid \gamma$ is a DNF formula equivalent to $\Sigma \mid \gamma$. Now, our claim is that the formula $\Sigma*$ obtained by keeping only the logically weakest terms $\gamma_i \mid \gamma$ among $(\bigvee_{i=1}^{n} \gamma_i) \mid \gamma$ is a prime implicants formula equivalent to $\Sigma \mid \gamma$. Removing such terms clearly is truth-preserving. Since generating $\Sigma*$ requires only $\mathcal{O}(n^2)$ entailment tests among terms, and since such tests can be easily achieved in polynomial time, we obtain that IP satisfies **CD**. Now, how to prove that $\Sigma*$ is in prime implicants form? Since any pair of different terms of $\Sigma*$ cannot be compared w.r.t. logical entailment, the correctness of Quine's consensus algorithm for generating prime implicants shows that it is sufficient to prove that every consensus among two terms of $\Sigma*$ is inconsistent or entails another term of $\Sigma*$. Let's recall that consensus is to DNF formulas what resolution is to CNF formulas. Since $\Sigma$ is in prime implicants form, every consensus among two terms of $\Sigma$ is inconsistent or entails another term of $\Sigma$. What happens to the terms (here, the prime implicants) of $\Sigma$ when conditioned by $\gamma$? All those containing the negation of a literal of $\gamma$ are removed and the remaining ones are shortened by removing from them every literal of $\gamma$. Hence, for every pair of terms $\gamma_1$, $\gamma_2$ of $\Sigma$, if there is no consensus between $\gamma_1$ and $\gamma_2$, then there is no consensus between $\gamma_1|\gamma$ and $\gamma_2|\gamma$: conditioning cannot create new consensus. Now, it remains to prove that no unproductive consensus between terms of $\Sigma$ can be rendered productive through conditioning. Formally, let $\gamma_1 = \gamma_1' \wedge l$ and $\gamma_2 = \gamma_2' \wedge \neg l$ be two prime implicates of $\Sigma$ s.t. $l$ (resp. $\neg l$) does not appear in $\gamma_1'$ (resp. $\gamma_2'$). There is a consensus $\gamma_1' \wedge \gamma_2'$ between $\gamma_1$ and $\gamma_2$. Let us assume that both $\gamma_1$ and $\gamma_2$ have survived the conditioning: this means that both $\gamma_1|\gamma$ and $\gamma_2|\gamma$ are consistent. Especially, $l$ belongs to $\gamma_1|\gamma$ and $\neg l$ belongs to $\gamma_2|\gamma$. Accordingly, there is a consensus between $\gamma_1|\gamma$ and $\gamma_2|\gamma$. By construction, this consensus is equivalent to $(\gamma_1'|\gamma) \wedge (\gamma_2'|\gamma)$, hence equivalent to $(\gamma_1' \wedge \gamma_2')|\gamma$. Now, if $\gamma_1' \wedge \gamma_2'$ is inconsistent, then $(\gamma_1' \wedge \gamma_2')|\gamma$ is inconsistent as well and we are done. Otherwise, let us assume that there exists a prime implicant $\gamma_3$ of $\Sigma$ s.t. $\gamma_1' \wedge \gamma_2' \models \gamma_3$ holds. Necessarily, $\gamma_3$ is preserved by the conditioning of $\Sigma$ by $\gamma$. Otherwise, $\gamma_3$ would contain the negation of a literal of $\gamma$, but since every literal of $\gamma_3$ is a literal of $\gamma_1$ or a literal of $\gamma_2$, $\gamma_2$ and $\gamma_3$ would not have both survived the conditioning. Since $\gamma_1' \wedge \gamma_2' \models \gamma_3$ holds, we necessarily have $(\gamma_1' \wedge \gamma_2')|\gamma \models \gamma_3|\gamma$. This completes the proof.
- MODS. Direct consequence of Lemma A.12 and the fact that MODS satisfies $\wedge$**BC** and **FO** (see below).

- **FO**.
  - DNNF and DNF. It is known that DNNF satisfies **FO** (Darwiche, 2001a). It is also known that DNF satisfies **FO** (Lang et al., 2000).
  - NNF, s-NNF, f-NNF, d-NNF, BDD and CNF. Let $\Sigma$ be a sentence in CNF. We now show that if any of the previous languages satisfies **FO**, then we can test the consistency of $\Sigma$ in polytime. Since CNF does not satisfy **CO** (see Proposition 4.1), it then follows that none of the previous languages satisfy **FO** unless P = NP. First, we note that $\Sigma$ must also belong to NNF and f-NNF. Moreover, $\Sigma$ can be turned into a sentence in BDD in polytime (Lemma A.8) or a sentence in s-NNF in polytime (see the proof of Lemma A.1). We also have that $\Sigma$ can be turned into a sentence in d-NNF in polytime since BDD $\subseteq$ d-NNF. Suppose now that one of the previous languages, call it **L**, satisfy **FO**. We can test the consistency of $\Sigma$ in polytime as follows:
    * Convert $\Sigma$ into a sentence $\Sigma*$ in **L** in polytime (as shown above).
    * Compute $\exists\, Vars(\Sigma*).\Sigma*$, which can be done in polytime by assumption.

* Test the validity of $\exists\, Vars(\Sigma*).\Sigma*$, which can be done in polytime since the sentence contains no variables—all we have to do is check whether the sentence evaluates to *true*.

  Finally, note that the definition of forgetting implies that a sentence $\Gamma$ is consistent iff $\exists\, Vars(\Gamma).\Gamma$ is valid, which completes the proof.

- d-DNNF and sd-DNNF. Follows immediately since none of these languages satisfies **SFO** unless P = NP (see below).
- IP. Follows immediately since IP does not satisfy **SFO**.
- FBDD, OBDD and $\text{OBDD}_<$. We will show that if FBDD (resp. OBDD, $\text{OBDD}_<$) satisfies **FO**, then for every sentence $\Gamma$ in DNF, there must exist an equivalent sentence $\Sigma$ in FBDD (resp. OBDD, $\text{OBDD}_<$), which size is polynomial in the size of $\Gamma$. This contradicts the fact that FBDD (resp. OBDD, $\text{OBDD}_<$) $\not\leq$ DNF—see Table 3.

  Given a DNF $\Gamma$ consisting of terms $\gamma_1$, ..., $\gamma_n$, we can convert each of these terms into equivalent FBDD (resp. OBDD, $\text{OBDD}_<$) sentences $\alpha_1, \ldots, \alpha_n$ in polytime. Let $\{v_1, \ldots, v_{n-1}\}$ be a set of variables that do not belong to $PS$. Construct a new set of variables $PS' = PS \cup \{v_1, \ldots, v_{n-1}\}$. In case of OBDD and $\text{OBDD}_<$, we also assume that these new variables are earlier than variables $PS$ in the ordering. Consider now the sentence $\Sigma = \exists\{v_1, \ldots, v_{n-1}\}.\Delta^1$, with respect to variables $PS'$, where $\Delta^i$ is inductively defined by:

  * $\Delta^i = \alpha_i$, for $i = n$, and
  * $\Delta^i = (\alpha_i \wedge v_i) \vee (\Delta^{i+1} \wedge \neg v_i)$, for $i = 1, \ldots, n-1$.

  Clearly enough, an FBDD (resp. OBDD, $\text{OBDD}_<$) sentence equivalent to $\Delta^1$ can be computed in time polynomial in the input size. Moreover, we have $\Sigma \equiv \bigvee_{i=1}^{n} \alpha_i \equiv \bigvee_{i=1}^{n} \gamma_i \equiv \Gamma$. Hence, if FBDD (resp. OBDD, $\text{OBDD}_<$) satisfies **FO**, then we can convert the DNF sentence $\Gamma$ into an equivalent FBDD (resp. OBDD, $\text{OBDD}_<$) which size is polynomial in the size of the given DNF. This is impossible in general.
- PI. It is known that the prime implicates of $\exists X.\Sigma$ are exactly the prime implicates of $\Sigma$ that do not contain any variable from $X$ (see Proposition 55 in (Marquis, 2000)). Hence, such prime implicates can be computed in time polynomial in the input size when $\Sigma$ is in prime implicates form.
- MODS. Given a MODS formula $\Sigma$ and a subset $X$ of $PS$, the formula obtained by removing every leaf node (and the corresponding incoming edges) of $\Sigma$ labeled by a literal $x$ or $\neg x$ s.t. $x \in X$ is a MODS representation of $\exists X.\Sigma$—this is an easy consequence of Propositions 18 and 20 from (Lang et al., 2000). See also the polytime operation of forgetting on DNNF, as defined in (Darwiche, 2001a), which applies to MODS, since MODS $\subseteq$ DNNF, and which can be easily modified so it guarantees that the output is in MODS when the input is also in MODS.

- **SFO**.
  - DNNF, DNF, PI and MODS. Immediate from the fact that each of these languages satisfies **FO** (see above).
  - NNF, d-NNF, s-NNF, f-NNF, BDD, $\text{OBDD}_<$ and CNF. Direct from the fact that $\exists x.\Sigma \equiv (\Sigma|x) \vee (\Sigma|\neg x)$ holds and the fact that any of these fragments satisfies **CD** and $\vee$**BC**.
  - OBDD. Direct from the fact that only one OBDD sentence is considered in the transformation and $\text{OBDD}_<$ satisfies **SFO**.
  - d-DNNF, sd-DNNF and FBDD. Let $\alpha_1$ and $\alpha_2$ be two FBDD formulas. Let $x$ be a variable not included in $Vars(\alpha_1) \cup Vars(\alpha_2)$. The formula $\Sigma = (x \wedge \alpha_1) \vee (\neg x \wedge \alpha_2)$ is a FBDD formula since decomposability and decision are preserved by this construction. Since $\exists x.\Sigma$ is equivalent to $\alpha_1 \vee \alpha_2$, if FBDD would satisfy **SFO**, it would satisfy $\vee$**BC** as well, but this is not the case unless P = NP (see below). The same conclusion can be drawn for d-DNNF. Hence, FBDD and d-DNNF do not satisfy **SFO** unless P = NP. Since every d-DNNF formula can be turned in polynomial time into an equivalent sd-DNNF formula, we obtain that sd-DNNF does not satisfy **SFO** unless P = NP.

– IP. Let us show that the number of prime implicants of $\exists x.\Sigma$ can be exponentially greater than the number of prime implicants of $\Sigma$. Let $\Sigma'$ be the following DNF formula:

$$\Sigma' = \left(\bigvee_{i=1}^{k} \bigvee_{j=1}^{m} (p_i \wedge q_{i,j})\right) \vee \bigwedge_{i=1}^{k} \neg p_i.$$

$\Sigma'$ has $(m+1)^k + mk$ primes implicants (Chandra & Markowsky, 1978). Now, let $\Sigma$ be the formula:

$$\Sigma = \left(\bigvee_{i=1}^{k} \bigvee_{j=1}^{m} (x \wedge p_i \wedge q_{i,j})\right) \vee (\neg x \wedge \bigwedge_{i=1}^{k} \neg p_i).$$

Since $\Sigma'$ can be obtained from $\Sigma$ by removing in every term of $\Sigma$ every occurrence of $x$ and $\neg x$, $\Sigma'$ is equivalent to $\exists\{x\}.\Sigma$ (see (Lang et al., 2000)). Now, $\Sigma$ has only $mk+1$ prime implicants; indeed, every term of it is a prime implicant, and the converse holds since every term is maximal w.r.t. logical entailment and every consensus of two terms is inconsistent. This completes the proof.

- $\wedge$**C**.

  – NNF, s-NNF, d-NNF, CNF. The property is trivially satisfied by these languages since determinism and smoothness are only concerned with or-nodes. Hence, if $\alpha_1, \ldots, \alpha_n$ belong to one of these languages, so is $\alpha_1 \wedge \ldots \wedge \alpha_n$.

  – BDD. It is well–known that the conjunction of two BDDs $\alpha$ and $\beta$ can be easily computed by connecting the 1-sink of $\alpha$ to the root of $\beta$ (see proof of Lemma A.8). The size of the resulting BDD is just the *sum* of the sizes of the respective BDDs of $\alpha$ and $\beta$. Accordingly, we can repeat this operation $n$ times in time polynomial in the input size.

  – f-NNF. Direct from the fact that f-NNF does not satisfy $\wedge$**BC**.

  – FBDD, OBDD, $\text{OBDD}_<$, DNF, PI and IP. It is straightforward to convert a clause into an equivalent formula in any of these languages in polynomial time. In the proof of Proposition 3.1, we show specific CNF formulas which cannot be turned into an equivalent FBDD (resp. OBDD, $\text{OBDD}_<$, DNF, PI and IP) formulas in polynomial space (see Tables 9 and 10). Hence, such conversion cannot be accomplished in polynomial time either. This implies that none of FBDD, OBDD, $\text{OBDD}_<$, DNF, PI and IP satisfies $\wedge$**C**.

  – DNNF, d-DNNF and sd-DNNF. Direct from the fact that none of these languages satisfy $\wedge$**BC** unless
  P = NP.

  – MODS. Let $\Sigma = \bigwedge_{i=1}^{n} \Sigma_i$, where $\Sigma_i = (x_{i,1} \vee x_{i,2})$, $i \in 1..n$. Each $\Sigma_i$ has 3 models over $Vars(\Sigma_i)$. Since $\Sigma$ has $3^n$ models, it does not have a MODS representation of size polynomial in the input size.

- $\wedge$**BC**.

- NNF, s-NNF, d-NNF, BDD and CNF. Immediate since each of these languages satisfy $\wedge$**C** (see above).
- DNNF, d-DNNF, sd-DNNF, FBDD and OBDD. Checking whether the conjunction of two $\text{OBDD}_<$ formulas $\alpha_1$ and $\alpha_2$ (w.r.t. two different variable orderings $<$) is consistent is NP-complete (see Lemma 8.14 in (Meinel & Theobald, 1998)). Since OBDD satisfies **CO**, it cannot satisfy $\wedge$**BC** unless P $=$ NP. Since OBDD $\subseteq$ FBDD $\subseteq$ d-DNNF $\subseteq$ DNNF, and d-DNNF and DNNF satisfy **CO**, none of them can satisfy $\wedge$**BC** unless P $=$ NP. Finally, since every d-DNNF formula can be turned in polynomial time into an equivalent smoothed d-DNNF formula and since sd-DNNF satisfies **CO**, it cannot be the case that sd-DNNF satisfy $\wedge$**BC** unless P $=$ NP.
- $\text{OBDD}_<$. Well-known fact (Bryant, 1986).
- f-NNF. Let $\alpha_1 = \bigwedge_{i=0}^{n-1}(x_{2i} \vee x_{2i+1})$ be a CNF formula and $\alpha_2 = \bigvee_{i=0}^{n-1}(x'_{2i} \wedge x'_{2i+1})$ a DNF formula. $\alpha_1$ has $2^n$ essential prime implicants and $n$ essential prime implicates (see the proof of Proposition 3.1, Table 9). By duality, $\alpha_2$ has $n$ essential prime implicants and $2^n$ essential prime implicates. Now, $\alpha_1$ and $\alpha_2$ are two f-NNF formulas. By Lemma A.13, we know that every f-NNF formula $\beta$ can be turned in polynomial time into a CNF formula or a DNF formula. If f-NNF would satisfy $\wedge$**BC**, then a f-NNF formula $\beta$ s.t. $\beta \equiv \alpha_1 \wedge \alpha_2$ could be computed in time polynomial in the input size. Hence, either a CNF formula equivalent to $\alpha_1 \wedge \alpha_2$ or a DNF formula equivalent to $\alpha_1 \wedge \alpha_2$ could be computed in polytime. But this is impossible since $\alpha_1 \wedge \alpha_2$ has $n + 2^n$ essential prime implicates and $n * 2^n$ essential prime implicants. Hence every CNF (resp. DNF ) formula equivalent to $\alpha_1 \wedge \alpha_2$ has a size exponential in $|\alpha_1| + |\alpha_2|$.

  Note that in the case where the two f-NNF formulas $\alpha_1$ and $\alpha_2$ into consideration can be turned in polynomial time into either two CNF formulas or two DNF formulas, then a f-NNF formula equivalent to $\alpha_1 \wedge \alpha_2$ can be computed in time polynomial in the input size (this is obvious when two CNF formulas are considered and the next item of the proof shows how this can be achieved when two DNF formulas are considered).
- DNF and MODS. If $\alpha_1$ and $\alpha_2$ are sentences in one of these languages **L**, then we can construct a sentence in **L** which is equivalent to $\alpha_1 \wedge \alpha_2$ by simply taking all the conjunctions of one term from $\alpha_1$ and one term from $\alpha_2$, while removing redundant literals in the resulting terms and removing any inconsistent terms in the result. The disjunction of all the resulting terms is a sentence from **L** equivalent to $\alpha_1 \wedge \alpha_2$ and it has been computed in polynomial time.
- PI. Let $\alpha_1 = \bigvee_{i=1}^{k} p_i$ and $\alpha_2 = \bigwedge_{i=1}^{k} \bigwedge_{j=1}^{m} (\neg p_i \vee q_{i,j})$. Sentence $\alpha_1$ has one prime implicate and $\alpha_2$ has $m * k$ prime implicates. But $\alpha_1 \wedge \alpha_2$ has $(m+1)^k + m * k$ prime implicates (Chandra & Markowsky, 1978).
- IP. Let $IP(\alpha)$ be the set of prime implicants for $\alpha$. We have $IP(\alpha_1 \wedge \alpha_2) = \max(\{\beta_1 \wedge \beta_2 \mid \beta_1 \in IP(\alpha_1), \beta_2 \in IP(\alpha_2)\}, \models)$ (up to logical equivalence). See e.g., (dual of) Proposition 40 in (Marquis, 2000).

- $\vee$**C**.
  - NNF, s-NNF, DNNF and DNF. The property is trivially satisfied by these languages since decomposability is only concerned with and-nodes, and since every NNF formula can be turned in polynomial time into an equivalent smoothed NNF formula.
  - d-NNF and BDD. Direct consequence from the fact that d-NNF and BDD satisfies both $\wedge$**C** and $\neg$**C**. Especially, it is well-known that the disjunction of two BDDs $\alpha$ and $\beta$ can be easily computed by connecting the 0-sink of $\alpha$ to the root of $\beta$ (see the proof of Lemma A.8). The size of the resulting BDD is just the *sum* of the sizes of the respective BDDs of $\alpha$ and $\beta$. Accordingly, we can repeat this operation $n$ times in time polynomial in the input size.
  - f-NNF. Since f-NNF does not satisfy $\wedge$**C** but satisfies $\neg$**C**, it cannot satisfy $\vee$**C** (due to De Morgan's laws).
  - FBDD, OBDD, $\text{OBDD}_<$, CNF, PI, IP and MODS. It is straightforward to convert any term into an equivalent formula from any of the previous languages in polynomial time. In the proof of Proposition 3.1, we show specific DNF formulas which cannot be turned into equivalent FBDD (resp. OBDD, $\text{OBDD}_<$, CNF , PI, IP and MODS) formulas in polynomial space (see Tables 9, 10 and 15). Hence, the conversion cannot be accomplished in polynomial time either. This implies that none of FBDD, OBDD, $\text{OBDD}_<$, CNF, PI, IP and MODS satisfies $\vee$**C**.
  - d-DNNF and sd-DNNF. Immediate form the fact that none of these classes satisfies $\vee$**BC** unless $\mathsf{P} = \mathsf{NP}$ (see below).
- $\vee$**BC**.
  - NNF, d-NNF, DNNF, s-NNF, BDD and DNF. Immediate since each of these languages satisfies $\vee$**C**.
  - $\text{OBDD}_<$. Well-known fact (Bryant, 1986).
  - OBDD, FBDD, d-DNNF and sd-DNNF. Checking whether the conjunction of two $\text{OBDD}_<$ formulas $\alpha_1$ and $\alpha_2$ (w.r.t. two different variable orderings $<$) is consistent is $\mathsf{NP}$-complete (see Lemma 8.14 in (Meinel & Theobald, 1998)). Now, $\alpha_1 \wedge \alpha_2$ is inconsistent iff $\neg\alpha_1 \vee \neg\alpha_2$ is valid. Since OBDD satisfies $\neg$**C**, an OBDD formula equivalent to $\neg\alpha_1$ (resp. $\neg\alpha_2$) can be computed in time polynomial in $|\alpha_1|$ (resp. $|\alpha_2|$). Since OBDD $\subseteq$ FBDD $\subseteq$ d-DNNF, the resulting formulas are also FBDD and d-DNNF formulas. If OBDD (resp. FBDD, d-DNNF) would satisfy $\vee$**BC**, then an OBDD (resp. FBDD, d-DNNF) formula equivalent to $\neg\alpha_1 \vee \neg\alpha_2$ could be computed in time polynomial in $|\alpha_1| + |\alpha_2|$. But since d-DNNF satisfies **VA**, this is impossible unless $\mathsf{P} = \mathsf{NP}$. Finally, since every d-DNNF formula can be turned in polynomial time into an equivalent sd-DNNF formula, sd-DNNF cannot satisfy $\vee$**BC** unless $\mathsf{P} = \mathsf{NP}$.
  - f-NNF. Since f-NNF does not satisfy $\wedge$**BC** but satisfies $\neg$**C**, it cannot satisfy $\vee$**BC** (due to De Morgan's laws).
  - CNF. If $\alpha_1$ and $\alpha_2$ are two CNF sentences, then we can construct a CNF sentence which is equivalent to $\alpha_1 \vee \alpha_2$ by simply taking all the disjunctions of one clause from $\alpha_1$ and one clause from $\alpha_2$, while removing redundant literals inside the resulting clauses and removing any valid clause in the result. The conjunction of all the resulting clauses is a CNF sentence equivalent to $\alpha_1 \vee \alpha_2$, and it has been computed in polynomial time.
  - PI. Let $PI(\alpha)$ be the set of prime implicates for sentence $\alpha$. We have $PI(\alpha_1 \vee \alpha_2) = \min(\{\beta_1 \vee \beta_2 \mid \beta_1 \in PI(\alpha_1), \beta_2 \in PI(\alpha_2)\}, \models)$. See Proposition 40 in (Marquis, 2000).

– IP. Let $\alpha_1 = \bigwedge_{i=1}^{k} p_i$ and $\alpha_2 = \bigvee_{i=1}^{k} \bigvee_{j=1}^{m} (\neg p_i \wedge q_{i,j})$. Sentence $\alpha_1$ has one prime implicant and $\alpha_2$ has $m * k$ prime implicants. But $\alpha_1 \vee \alpha_2$ has $(m+1)^k + m * k$ prime implicants (Chandra & Markowsky, 1978).

– MODS. Let $\alpha_1 = \bigwedge_{i=1}^{n} x_i$ and $\alpha_2 = y$. Sentence $\alpha_1$ has 1 model over $Vars(\alpha_1)$ and $\alpha_2$ has 1 model over $Vars(\alpha_2)$. But $\alpha_1 \vee \alpha_2$ has $2^n + 1$ models over $Vars(\alpha_1) \cup Vars(\alpha_2)$.

- $\neg$**C**.
  - NNF, s-NNF, f-NNF, BDD, FBDD, OBDD and $\text{OBDD}_<$. The property is obviously satisfied by NNF. s-NNF also satisfies $\neg$**C** since every NNF formula can be turned in polynomial time into an equivalent s-NNF formula. f-NNF satisfies $\neg$**C** since applying De Morgan's laws on a f-NNF formula results in a f-NNF formula. Finally, for all the forms of BDDs, it is sufficient to switch the labels of the sinks to achieve negation (Bryant, 1986).
  - CNF. Because the negation of a DNF formula is a CNF formula that can be computed in polynomial time, if CNF would satisfy $\neg$**C**, then it would be possible to turn any DNF formula into an equivalent CNF formula in polynomial time (by involution of negation). But we know that it is not possible in polynomial space since CNF $\not\leq$ DNF(see the proof of Proposition 3.1). Hence, CNF does not satisfy $\neg$**C**.
  - DNF. Dual of the proof just above (just replace CNF by DNF and *vice-versa*).
  - PI. The formula $\Sigma_n = \bigwedge_{i=0}^{n-1}(x_{2i} \vee x_{2i+1})$ is in prime implicates form (see the proof of Proposition 3.1, Table 9). This formula has exponentially many prime implicants, that are just the negations of the prime implicates of $\neg\Sigma_n$. Since $\neg\Sigma_n$ has exponentially many prime implicates, it cannot be the case that PI satisfies $\neg$**C**.
  - IP. We just have to take the dual of the above proof (prime implicates case). The formula $\Sigma_n = \bigvee_{i=0}^{n-1}(x_{2i} \wedge x_{2i+1})$ is in prime implicants form. This formula has exponentially many prime implicates, that are just the negations of the prime implicants of $\neg\Sigma_n$. Since $\neg\Sigma_n$ has exponentially many prime implicants, it cannot be the case that IP satisfies $\neg$**C**.
  - DNNF. The negation of any CNF formula can be computed in polynomial time as a DNF formula, hence as a DNNF formula. If DNNF would satisfy $\neg$**C**, then it would be possible to turn a CNF formula into an equivalent DNNF one (by involution of negation). Because DNNF satisfies **CO**, we would have P = NP.
  - d-NNF. Following is a procedure for negating a d-NNF sentence $\Delta$:[9]
    - ∗ Traverse nodes in the DAG of $\Delta$, visiting the children of a node before you visit the node itself. When visiting a node, construct its negation as follows:
      - · *true* is the negation of *false*.
      - · *false* is the negation of *true*.
      - · $\wedge(N'_1, \ldots, N'_k)$ is the negation of $\vee(N_1, \ldots, N_k)$. Here, $N'_i$ is the node representing the negation of $N_i$.
      - · $\vee(\wedge(N'_1, M_1), \ldots, \wedge(N'_k, M_k))$ is the negation of $\wedge(N_1, \ldots, N_k)$. Here, $N'_i$ is the node representing the negation of $N_i$, and $M_i$ is a node representing the conjunction $N_1 \wedge \ldots \wedge N_{i-1}$.
    - ∗ Return the negation of the root of d-NNF $\Delta$.

    We can implement the above four steps so that we when we visit a node with $k$ children, we only construct $O(k)$ nodes and $O(k)$ edges.[10] Hence, the procedure complexity is

9. Mark Hopkins pointed us to this procedure.
10. We assume that any or-node (resp. and-node) with less than two children is removed and replaced by its unique child or by *false* (resp. *true*) if it has no children. This simplification process is equivalence-preserving and it can be achieved in time linear in the size of the input DAG.

linear in the size of the original d-NNF. It is easy to check that the result is equivalent to the negation of the given d-NNF sentence and is also in d-NNF.

- sd-DNNF and d-DNNF. Unknown.
- MODS. $\Sigma = \bigwedge_{i=1}^{n} x_i$ has only one model over $\bigcup_{i=1}^{n}\{x_i\}$ but its negation $\neg\Sigma$ has $2^n - 1$ models over $\bigcup_{i=1}^{n}\{x_i\}$. Hence MODS cannot satisfy $\neg$**C**. □